\pdfoutput=1
\documentclass[11pt]{article}

\usepackage[]{EACL2023}

\usepackage{times}
\usepackage{latexsym}

\usepackage{nicematrix}
\usepackage[T1]{fontenc}
\usepackage[utf8]{inputenc}

\usepackage{microtype}

\usepackage{inconsolata}

\title{A Methodology for Generative Spelling Correction via Natural Spelling Errors Emulation across Multiple Domains and Languages}

\author{Nikita Martynov \\
  SberDevices / Moscow \\
  \texttt{nikita.martynov.98@list.ru} \And 
  Mark Baushenko\\
  SberDevices / Moscow \\
  \texttt{m.baushenko@gmail.com} \And 
  Anastasia Kozlova \\
  SberDevices / Moscow \\
  \texttt{anastasi2510@gmail.com} \\\AND
  Katerina Kolomeytseva \\
  SberDevices / Moscow \\
  \texttt{kolomeytsevak@gmail.com} \And 
  Aleksandr Abramov \\
  SberDevices / Moscow \\
  \texttt{andril772@gmail.com} \And 
  Alena Fenogenova \\
  SberDevices / Moscow \\
  \texttt{alenush93@gmail.com}
}

\begin{document}
\maketitle
\begin{abstract}
Modern large language models demonstrate impressive capabilities in text generation and generalization. However, they often struggle with solving text editing tasks, particularly when it comes to correcting spelling errors and mistypings.
In this paper, we present a methodology for generative spelling correction (SC), which was tested on English and Russian languages and potentially can be extended to any language with minor changes.
Our research mainly focuses on exploring natural spelling errors and mistypings in texts and studying the ways those errors can be emulated in correct sentences to effectively enrich generative models' pre-train procedure. 
We investigate the impact of such emulations and the models' abilities across different text domains. In this work, we investigate two spelling corruption techniques: 1) first one mimics human behavior when making a mistake through leveraging statistics of errors from particular dataset and 2) second adds the most common spelling errors, keyboard miss clicks, and some heuristics within the texts.
We conducted experiments employing various corruption strategies, models' architectures and sizes on the pre-training and fine-tuning stages and evaluated the models using single-domain and multi-domain test sets. As a practical outcome of our work, we introduce SAGE~\footnote{\url{https://github.com/ai-forever/sage/}} (Spell checking via Augmentation and  Generative distribution Emulation). It is a library for automatic generative SC that includes a family of pre-trained generative models and built-in augmentation algorithms. 
\end{abstract}

\section{Introduction}
\label{sec:intro}

Recent advancements in large language models have shown remarkable capabilities in text generation and language understanding that can be seen on various benchmarks such as SuperGLUE~\cite{wang2020superglue}, GEM~\cite{gehrmann2021gem}, BigBench~\cite{srivastava2023beyond} etc. However, these models often encounter challenges when it comes to effectively addressing text editing tasks, particularly automatic correction of misspelling and mistyping. The task is well known, and many traditional approaches rely on explicit rules, dictionaries, or statistical models to detect and correct spelling errors. However, the emergence of large language models and generative techniques has introduced new possibilities and improved the effectiveness of automatic spelling correction (SC).

Thus, in this paper, we address the task of automatic generative SC across various domains. Our research primarily studies natural orthographic errors, text misspellings, and their emulation during model pre-training. We explore the impact of these emulations on the model's abilities across different domains and models.

As part of our methodology, we leverage two different spelling corruption techniques. The first technique applies the statistical analysis of common errors, aiming to mimic natural human behavior when making mistakes. The second technique introduces the most frequent spelling errors, keyboard miss clicks, and a set of heuristics within the texts. We conduct experiments for the Russian and English languages with various corruption strategies and model sizes during the pre-training and fine-tuning stages. As our work's practical result, we introduce SAGE (Spellchecking via Augmentation and  Generative distribution Emulation) — a comprehensive library for automatic generative SC that incorporates a range of generative models, trained using our proposed methodology, and offers built-in augmentation techniques. Additionally, SAGE contains the data hub, a valuable resource for the Russian language, consisting of novel spelling datasets.

The remainder is structured as follows. We overview multiple prior works on SC and augmentation strategies for data corruption in Section~\ref{sec:related_work}. 
Section~\ref{sec:method} presents our methodology, including task formulation, methodology overview, the precise approaches of the corruption techniques, and the data we used. Section~\ref{sec:experiments} lists the experiments and the generative models we used and demonstrates the effectiveness of our proposed techniques and the impact of different model configurations. We report the achieved results in Section~\ref{sec:evaluation} and analyze the obtained scores. Section~\ref{sec:conclusion} concludes with a discussion of the future work directions.

\section{Related work}
\label{sec:related_work}

Spell checking is a fundamental task in natural language processing (NLP) that aims to correct misspelled words in text automatically. Multiple approaches have been proposed to tackle this task, namely rule-based, statistical, and generative SC methods, which will be examined in this section. 

Rule-based spell checking is one of the most common approaches that relies on predefined rules and dictionaries for detecting and rectifying misspelled words. These resources can incorporate algorithmic error models such as Longest Common Subsequence ~\cite{taghva2001ocrspell}, Levenshtein Distance ~\cite{van2004supervised}, or Phonetic Algorithms ~\cite{kondrak2006evaluation}.

Statistical spell checking approaches employ machine learning algorithms to learn from extensive text corpora. These algorithms can identify common spelling errors and their corresponding corrections. Some examples of statistical approaches include n-gram models~\cite{ahmed2009revised}, Hidden Markov Models~\cite{stuker2011towards}, part-of-speech tagging~\cite{vilares2016studying} and Noisy Channel Model~\cite{kernighan1990spelling}.

Generative SC is a novel spell checking approach that has shown promising results in recent years. Such systems take into account the context, due to the architecture nature of language models such as seq2seq Long Short-Term Memory (LSTM)~\cite{evershed2014correcting}, seq2seq Bidirectional LSTM~\cite{zhou2017spelling}, and state-of-the-art transformer models like BERT~\cite{sun2019contextual}, BSpell~\cite{rahman2022bspell}, etc.

The paper~\cite{guo2019zero} presents multilingual translation models for paraphrase generation task.  M2M100 models~\cite{DBLP:journals/corr/abs-2010-11125} (Many-to-Many multilingual models) effectively translate source language text into a target language that aligns with the source language. Given the M2M100 models' comprehensive understanding of multiple languages, their utilization in spell checking tasks proves promising. In our research, among other investigations, we explore the suitability of the M2M approach for spell checking.

\textit{Datasets}. English spell checking research has received significant attention due to English widespread use, which results in the creation of spell checking datasets. Evaluation datasets such as BEA-2019 shared task~\cite{bryant2019bea}, comprising corpora like FCE~\cite{yannakoudakis2011new}, W\&I+LOCNESS, Lang-8~\cite{tajiri2012tense}, and NUCLE~\cite{dahlmeier2013building}, provide valuable resources for assessing spell checking and error correction tasks. NeuSpell \cite{jayanthi2020neuspell} introduced the BEA60K natural test set and the well-established JFLEG dataset~\cite{napoles2017jfleg}, containing only spelling mistakes. Other clean corpora, including the Leipzig Corpora Collection~\cite{biemann2007leipzig} and the Gutenberg corpus~\cite{gerlach2020standardized}, offer diverse sources such as news, web content, and books for further exploration in spell checking research.

Among the standard open source datasets for the Russian language is RUSpellRU~\footnote{\url{https://www.dialog-21.ru/evaluation/2016/spelling\_correction/}}, which emerged after the competition on automatic SC for Russian social media texts~\cite{sorokin2016spellrueval}. Other open sources include the GitHub Typo Corpus~\cite{DBLP:journals/corr/abs-1911-12893}, which contains the Russian section, and the recent work~\cite{martynov2023augmentation}, which introduces a multi-domain dataset.

\textit{Text corruption methods}.
For training generative SC models, building a parallel corpus is essential. There are several ways to emulate spelling errors or augment the existing datasets. The example is the GEM benchmark and its associated augmentation library NL-Augmenter~\cite{dhole2023nl} and the work ~\cite{kuznetsov2021spelling} with the method for creating artificial typos. For the Russian language, the RuTransform framework~\cite{taktasheva-etal-2022-tape} presents adding noise into data through spelling corruption. Also, augmentation methods are proposed by ~\cite{martynov2023augmentation}.

\section{Methodology}
\label{sec:method}
In this work, we wanted our models to be built upon the criterion that meets the demands of their end users. The areas of potential utilization of SC tools abound with the language of varying orthographies and styles. Hence it imposes additional requirements for text editing systems. We decided to complement and, in some sense, complicate the straightforward paradigm of treating standard language as the only correct spelling option. In this section, we define the notion of SC task and describe our methodology in depth.

\subsection{Task Formalization}
\label{sec:formalization}

Before defining the SC task, we must establish the \textit{correct spelling} notion we employ in this work. Instead of rigorously normalizing all supposedly erroneous lexemes to the standard language, we propose distinguishing unintentional spelling violations from intentional ones. Plain language, colloquialisms, dialectisms, and abbreviations can express emotions and endow a text with distinct stylistic features. Since the act of intentional violation of spelling can hardly be expressed in terms of strict rules, it seems nearly impossible to distinguish intentional errors automatically. Instead, we use manual annotation as described in ~\cite{martynov2023augmentation}. Following ~\cite{martynov2023augmentation}, we consider a sentence annotated and emended by native experts as correct. Given a correct sentence, any sentence obtained from the correct one by (probably) multiple insertions, deletions, substitutions, or transpositions of characters is considered erroneous. This leads to the following definition of SC task that we use in this paper:

\begin{quote}
    Let $X=[x_1,...,x_N]=X_{corr.} \cup X_{incorr.}$, where $x_1,...,x_N$ is an ordered sequence of lexemes, $X_{corr.}=\{x_{i}\}_{i=1}^{k}$ is a set of correct lexemes, $X_{incorr.}=\{x_{j}\}_{j=1}^{p}$ is a set of incorrect lexemes, $p+k=N, p \ge 0, k > 0$, be the sentence that may contain spelling errors. The system $M$ then should produce corresponding sequence (ordered) $Y=[y_1,...,y_M]=Y_{corr.} \cup Y_{incorr.}, Y_{incorr.}=\emptyset$ so that
    \begin{enumerate}
        \item Correct lexemes are not modified: $!\exists f: \{x_{i}\}_{i=1}^{k} \rightarrow Y, f - $injective and preserves order and $f(x_i) = x_i$; 
        \item Original style of a sentence $X$ is preserved;
        \item All the information is fully transfered from $X$ to $Y$ and no new information appears in $Y$; 
    \end{enumerate}
\end{quote}

Basically, system $M$ only corrects unintentional errors and carry stylistic and factological pallet the same from $X$ to $Y$.

\subsection{Overview}

In this paper, we propose a methodology for generative SC, exploring the natural spelling errors across multiple domains and assessing their influence on spell-checking quality during pre-training and fine-tuning stages. The method can be summarized as follows:

\textbf{Corruption step}: the paper explores the methods of text corruption techniques using two augmentation methods. The first \textit{statistic-based approach} emulates the natural distribution of orthographic errors. The second \textit{heuristic-based} approach adds heuristics and related to it frequent noise to the data in some proportion without any given distribution of the particular domain parallel data set.

\textbf{Generation step}: we pre-train the generative models of different sizes and on the extensive synthetic dataset of diverse domains. The error distribution of the synthetic pre-train data is created by emulating the natural distribution of the errors via a statistic-based approach.

\textbf{Fine-tune step}: during the fine-tuning, we investigate the influence of corruption and domains on the final results. The models are evaluated on fixed single-domain and multiple-domain test sets. The experiments involve training the pre-trained models on various training data from single and multiple domains, as well as using the same data corrupted with the two aforementioned augmentation techniques.

The methodology is explored and tested in the Russian and English languages but can be potentially transferred to any language.

\subsection{Augmentations Strategies}
\label{section:augmentation}
We operate two strategies to introduce errors in sentences. This section provides a brief overview of those strategies. 

\subsubsection{Heuristic-based spelling corruption}
The first strategy represents spelling corruption through exploiting various heuristics, common error statistics, and understanding of implicit mechanics of a language. Nlpaug~\cite{ma2019nlpaug} and NeuSpell~\cite{jayanthi2020neuspell} libraries for English and Augmentex~\cite{martynov2023augmentation} for Russian are notable examples of such strategy. In this work, we choose Augmentex~\cite{martynov2023augmentation} for experiments with Russian language models. This library is accompanied with proven effectiveness for the Russian language ~\cite{martynov2023augmentation} and provides a flexible interface to its interior methods. Each method is responsible for modeling a specific type of error, including inserting random characters, replacing correctly spelled words with their incorrect counterparts, inserting nearby keyboard characters, and replacing a character with another based on the probability of its erroneous use. Augmentex allows researchers to control the distribution of error noise on word and sentence levels as well. In our experiments, we investigate Augmentex in depth by augmenting fine-tune datasets and studying its impact on models' performance. See details of its configurations used at the augmentation stage in~\ref{sec:appendix:augmentations}.

\subsubsection{Statistic-based spelling corruption}

We choose statistic-based spelling corruption (SBSC) from~\cite{martynov2023augmentation} as an attempt to reproduce errors from a particular piece of text. The method mimics human behavior when committing an error by scanning distributions of errors in a given text and then reapplying them on correct sentences. The algorithm requires a parallel corpus of sentence pairs (corrupted\_sentence, correct\_sentence): it builds a Levenshtein matrix between prefixes of sentences in each pair, then it traverses this matrix back along the main diagonal starting from the bottom right entry. At each step, the algorithm detects a position of an error in a sentence and its corresponding type based on surrounding entries. A detailed description of statistic-based spelling corruption is provided in~\cite{martynov2023augmentation}. Our work employs statistic-based spelling corruption to prepare pre-training datasets for both English and Russian generative models. We believe our research reveals SBSC's ability to be transferred to another language other than Russian, which it was initially proposed for in~\cite{martynov2023augmentation}. We also investigate the capacity of this noising strategy by experimenting with augmentation through spelling corruption while fine-tuning. 

\subsection{Datasets}
\label{sec:datasets}

For our multi-domain spell checking experiments, we developed three distinct data suites. 

\textbf{Golden Test Sets}: Fixed datasets, including both single-domain and multiple-domain texts, used for evaluation purposes.

\textbf{Pre-trained Data}: Synthetic data generated to emulate natural and random noise misspellings, employed during the pre-training stage to assess their impact on model performance.

\textbf{Training Data for fine-tuning}: Collected using the same method as the test sets, also corrupted with the proposed augmentation strategies to introduce diverse errors. Used during the fine-tuning stage to explore the impact of the different noise on the model performance across domains.

Below we describe the sets in detail.

\subsubsection{Golden Test Sets}

The datasets for the golden test set are chosen in accordance with the specified criteria. First, \textit{domain variation}: half of the datasets are chosen from different domains to ensure diversity, while the remaining half are from a single domain. This is done separately for English and Russian languages. Another criterion is \textit{spelling orthographic mistakes}: the datasets exclusively comprised mistyping, omitting grammatical or more complex errors of non-native speakers. This focus on spelling errors aligns with the formalization of the task as described in section~\ref{sec:formalization}.

For the Russian language, we choose four different sets:

\textbf{RUSpellRU} – the single-domain open source dataset for social media texts presented in the Shared Task~\cite{sorokin2016spellrueval}.

\textbf{MultidomainGold} – the dataset first presented in the paper~\cite{martynov2023augmentation}. It’s a multi-domain corpus comprising the domains: internet domain presented by the Aranea web-corpus, literature, news, social media, and strategic documents. We followed the methodological criteria of the paper and reproduced the two-stage annotation project via a crowd-sourcing platform Toloka~\footnote{\url{https://toloka.ai/tolokers}}: at the first stage, annotators are asked to correct the mistakes, on the second – to validate the results from the previous step. The statistics and details of the instructions and annotation schema are presented in the Appendix~\ref{sec:appendix:data} and~\ref{sec:appendix:annotation}. Following the annotation methodology, we extend the author's dataset with two more domains: reviews (the part of the Omnia set~\cite{pisarevskaya2022wikiomnia}) and subtitles (the part of the Russian part of the OpenSubtitles set~\footnote{\url{https://opus.nlpl.eu/OpenSubtitles-v2016.php}}).

\textbf{GitHubTypoCorpusRu} – we take the Russian part of the corpora introduced in work ~\cite{DBLP:journals/corr/abs-1911-12893}. Additionally, we validate the parallel data of this corpus by the same Toloka project, but only the second step from the methodology.

\textbf{MedSpellChecker}~\footnote{\url{https://github.com/DmitryPogrebnoy/MedSpellChecker/tree/main}} is a single-domain set of a specific lexicon of the medical domain; the multi-domain set above does not cover that. The set contains the medical texts of anamnesis. The data was verified via a two-stage annotation pipeline as well.

For the English language, we used two sets:
\textbf{BEA60K} is a multi-domain dataset corpus for spelling mistakes in English.

\textbf{JHU FLuency-Extended GUG Corpus
(JFLEG) dataset} is a single domain set, the spelling part. The dataset contains 2K spelling mistakes (6.1\% of all tokens) in 1601 sentences.

The test datasets statistics is presented in the Table~\ref{tab:testdata_statistics} of the Appendix, the annotation details in Appendix~\ref{sec:appendix:annotation}.

\subsubsection{Pre-training Data}
\label{section:pretraindata}

To prepare pre-training datasets, we take correct samples and then corrupt them employing augmentation strategies described in~\ref{section:augmentation}. As for correct samples for experiments in Russian, we use twelve gigabytes (12GB) of raw Russian Wikipedia dumps and an open source dataset of transcripted videos in Russian~\footnote{\url{https://huggingface.co/datasets/UrukHan/t5-russian-spell\_I}} of three and a half million (3.5M) texts. We remove all the sentences with characters other than Russian and English alphabets, digits, and punctuation or under forty characters. We balance both datasets to roughly 3.3 million sentences, resulting in a pre-training corpus of 6.611.990 texts. Then statistic-based spelling corruption is applied. We scan statistics from the train split of RUSpellRU ~\cite{sorokin2016spellrueval}, multiply the number of errors per sentence distribution by ten to ensure we induce a much denser noise in the pre-training corpus than it is in fine-tuning datasets, and apply to the pre-training corpus to get corrupted sentences. As a result, the pre-training dataset is a collection of 6.611.990 text pairs, each consisting of corrupted sentences and corresponding correct sentences. 

For pre-training in the English language, we combine clean Leipzig Corpora Collection~\footnote{\url{https://corpora.uni-leipzig.de}} (News domain) and English Wikipedia dumps, clean them the way we applied for Russian and create a parallel corpus using a statistic-based augmentation technique based on a 5k subset of BEA60K. We result in six gigabytes (6 GB) of data for pre-training.

\subsubsection{Training Data for fine-tuning}

As for the datasets for fine-tuning, we use train splits of RUSpellRU~\cite{sorokin2016spellrueval} and MultidomainGold~\cite{martynov2023augmentation} and a combination of both. You can see details in Table~\ref{tab:traindata_statistics}. We also employ spelling corruption methods from~\ref{section:augmentation} for augmentation purposes in two separate ways. First, we introduce misspellings in erroneous parts of train splits of fine-tuned datasets, inducing more errors without expanding the dataset itself. In the second strategy, we expand train splits of fine-tuned datasets. We obtain correct sentences from a particular dataset, corrupt spelling, and append pairs of corrupted sentences and corresponding correct sentences to the same dataset. In Tables~\ref{tab:results_all_augmentations} and ~\ref{tab:results_dev_set} first strategy is marked as \textit{Add} and the second as \textit{Concat}.

We do not prepare fine-tuned datasets for the English language since we do not conduct fine-tuning in our experiments. 

\section{Experiments}
\label{sec:experiments}
We conducted a comprehensive series of experiments involving diverse spelling corruption strategies over the encoder-decoder generative models of different sizes throughout the pre-training and fine-tuning phases as well as zero-shot evaluation of the pre-trained models. The models' statistics are presented in Table~\ref{tab:model_statistics}. We compared performance based on single-domain and multi-domain test sets. Furthermore, we conducted a comparative evaluation of the OpenAI models utilizing different prompts and standard open source models.

\begin{table*}[htbp!]
\centering
\footnotesize
\begin{NiceTabular}{{c}|{c}{c}{c}|{c}{c}{c}|{c}{c}{c}|*{3}{p{0.7cm}}}
\hline
\Block{2-1}{\textbf{Model}} & \Block{1-3}{\textbf{RUSpellRU}} & & &  \Block{1-3}{\textbf{MultidomainGold}} & & & \Block{1-3}{\textbf{MedSpellChecker}} & & & \Block{1-3}{\textbf{GitHubTypoCorpusRu}} \\ 
& Prec. & Rec. & F1 & Prec. & Rec. & F1 & Prec. & Rec. & F1 & Prec. & Rec. & F1 \\
\hline
\Block[l]{}{\textbf{M2M100-1.2B}} & & & & & & & & & & & & \\
%\multicolumn{1}{|r|}{Item3}
\Block[l]{}{\;\;\;\scriptsize{Pre-train (PT.)}} 
& 59.4 & 43.3 & 50.1
& 56.4 & 44.8 & 49.9
& 63.7 & 57.8 & 60.6
& 45.7 & 41.4 & 43.5 \\
\hline
\Block[l]{}{\;\;\;\scriptsize{RUSpellRU (+PT.)}} 
& 82.9 & \textbf{72.5} & 77.3
& 53.3 & 57.8 & 55.5
& 55.9 & 57.8 & 56.9
& 39.3 & 41.5 & 40.4 \\
\Block[l]{}{\;\;\;\scriptsize{RUSpellRU}} 
& 68.8 & 42.6 & 52.6
& 17.9 & 25.2 & 21.0
& 16.3 & 17.7 & 17.0
& 15.1 & 14.9 & 15.0 \\
\Block[l]{}{\;\;\;\scriptsize{MultidomainGold (+PT.)}} 
& 84.9 & 65.0 & 73.7
& 62.5 & 60.9 & 61.7
& 76.3 & \textbf{73.9} & \textbf{75.1}
& \textbf{47.9} & \textbf{43.3} & \textbf{45.5} \\
\Block[l]{}{\;\;\;\scriptsize{MultidomainGold}} 
& 75.4 & 35.7 & 48.5
& 46.5 & 39.9 & 43.0
& 69.1 & 31.0 & 42.8
& 27.4 & 18.6 & 22.1 \\
\Block[l]{}{\;\;\;\scriptsize{RUSpellRU+MDG (+PT.)}} 
& \textbf{88.8} & 71.5 & \textbf{79.2}
& \textbf{63.8} & \textbf{61.1} & \textbf{62.4}
& \textbf{78.8} & 71.4 & 74.9
& 47.1 & 42.9 & 44.9 \\
\Block[l]{}{\;\;\;\scriptsize{RUSpellRU+MDG}} 
& 81.2 & 47.4 & 59.9
& 45.8 & 37.0 & 40.9
& 71.8 & 39.1 & 50.7
& 26.1 & 17.4 & 20.9 \\
\hline

\Block[l]{}{\textbf{M2M100-418M}} & & & & & & & & & & & & \\
%\multicolumn{1}{|r|}{Item3}
\Block[l]{}{\;\;\;\scriptsize{Pre-train (PT.)}} 
& 57.7 & 61.2 & 59.4
& 32.8 & 56.3 & 41.5
& 23.2 & 64.5 & 34.1
& 27.5 & \textbf{42.6} & 33.4 \\
\hline
\Block[l]{}{\;\;\;\scriptsize{RUSpellRU (+PT.)}} 
& 81.8 & 63.4 & 71.4
& 45.3 & 55.9 & 50.0
& 40.8 & 52.2 & 45.8
& 29.5 & 36.6 & 32.7 \\
\Block[l]{}{\;\;\;\scriptsize{RUSpellRU}} 
& 66.5 & 38.5 & 48.8
& 20.9 & 26.0 & 23.2
& 22.3 & 14.8 & 17.8
& 11.4 & 13.2 & 12.2 \\
\Block[l]{}{\;\;\;\scriptsize{MultidomainGold (+PT.)}} 
& 81.3 & 55.4 & 65.9
& 57.9 & 56.5 & 57.2
& \textbf{73.5} & \textbf{66.0} & \textbf{69.5}
& 40.3 & 39.2 & 39.8 \\
\Block[l]{}{\;\;\;\scriptsize{MultidomainGold}} 
& 63.5 & 31.6 & 42.2
& 39.5 & 34.9 & 37.0
& 55.2 & 32.5 & 40.9
& 23.1 & 15.5 & 18.5 \\
\Block[l]{}{\;\;\;\scriptsize{RUSpellRU+MDG (+PT.)}} 
& \textbf{87.6} & \textbf{64.4} & \textbf{74.2}
& \textbf{60.3} & \textbf{56.6} & \textbf{58.4}
& 73.1 & 62.4 & 67.3
& \textbf{42.8} & 37.8 & \textbf{40.2} \\
\Block[l]{}{\;\;\;\scriptsize{RUSpellRU+MDG}} 
& 74.0 & 45.2 & 56.1
& 39.8 & 34.4 & 36.9
& 59.5 & 38.4 & 46.7
& 24.7 & 18.0 & 20.8 \\
\hline

\Block[l]{}{\textbf{FredT5-large}} & & & & & & & & & & & & \\
%\multicolumn{1}{|r|}{Item3}
\Block[l]{}{\;\;\;\scriptsize{Pre-train (PT.)}} 
& 58.5 & 42.4 & 49.2
& 42.5 & 42.0 & 42.2
& 37.2 & 51.7 & 43.3
& 52.7 & 41.7 & 46.6 \\
\hline
\Block[l]{}{\;\;\;\scriptsize{RUSpellRU (+PT.)}} 
& 55.1 & 73.2 & 62.9
& 26.7 & 55.1 & 36.0
& 12.9 & 49.6 & 20.4
& 26.2 & 40.5 & 31.8 \\
\Block[l]{}{\;\;\;\scriptsize{RUSpellRU}} 
& 40.7 & 50.4 & 45.0
& 20.5 & 42.4 & 27.6
& 6.9 & 26.0 & 11.0
& 15.2 & 23.8 & 18.6 \\
\Block[l]{}{\;\;\;\scriptsize{MultidomainGold (+PT.)}} 
& 67.7 & 60.2 & 63.8
& \textbf{61.7} & 60.5 & \textbf{61.1}
& 39.5 & \textbf{60.4} & \textbf{47.7}
& \textbf{69.3} & 44.6 & \textbf{54.3} \\
\Block[l]{}{\;\;\;\scriptsize{MultidomainGold}} 
& 49.6 & 39.9 & 44.2
& 48.1 & 43.4 & 45.6
& \textbf{43.2} & 41.2 & 42.2
& 50.8 & 25.7 & 34.1 \\
\Block[l]{}{\;\;\;\scriptsize{RUSpellRU+MDG (+PT.)}} 
& \textbf{74.5} & \textbf{73.4} & \textbf{73.9}
& 58.3 & \textbf{63.1} & 60.6
& 37.5 & 59.3 & 45.9
& 61.2 & \textbf{45.4} & 52.1\\
\Block[l]{}{\;\;\;\scriptsize{RUSpellRU+MDG}} 
& 56.3 & 56.2 & 56.3
& 48.2 & 48.5 & 48.3
& 42.5 & 42.7 & 42.6
& 49.4 & 26.9 & 34.8 \\

\end{NiceTabular}
\caption{The models' performance in experiments configurations for the Russian language. For each model, the experiments are reported for the pre-train model on zero-shot, the raw model fine-tuned on the specific train set, and the pre-train model ($+PT.$) fine-tuned on the specific train set. Metrics are reported in \textbf{Prec}ision / \textbf{Rec}all / \textbf{F1}-measure format from ~\cite{sorokin2016spellrueval}.}
\label{tab:results_rus_generative}
\end{table*}

\begin{table*}[htbp!]
\centering
\footnotesize
\begin{NiceTabular}{{c}|{c}{c}{c}|{c}{c}{c}|{c}{c}{c}|*{3}{p{0.7cm}}}
\hline
\Block{2-1}{\textbf{Model}} & \Block{1-3}{\textbf{RUSpellRU}} & & &  \Block{1-3}{\textbf{MultidomainGold}} & & & \Block{1-3}{\textbf{MedSpellChecker}} & & & \Block{1-3}{\textbf{GitHubTypoCorpusRu}} \\ 
& Prec. & Rec. & F1 & Prec. & Rec. & F1 & Prec. & Rec. & F1 & Prec. & Rec. & F1 \\
\hline
\Block[l]{}{Yandex.Speller}
& 83.0 & 59.8 & 69.5 
& 52.9 & 51.4 & 52.2
& \textbf{80.6} & 47.8 & 60.0
& \textbf{67.7} & 37.5 & 48.3 \\
\Block[l]{}{JamSpell}
& 42.1 & 32.8 & 36.9 
& 25.7 & 30.6 & 28.0
& 24.6 & 29.7 & 26.9
& 49.5 & 29.9 & 37.3 \\
\Block[l]{}{Hunspell}
& 31.3 & 34.9 & 33.0
& 16.2 & 40.1 & 23.0
& 10.3 & 40.2 & 16.4
& 28.5 & 30.7 & 29.6 \\
\hline
\Block[l]{}{gpt-3.5-turbo-0301} &  &  &  &  &  &  &  &  &  & & & \\
\Block[l]{}{\;\;\;\scriptsize{With Punctuation}} & 55.8 & 75.3 & 64.1 & 33.8 & 72.1 & 46.0 & 53.7 & 66.1 & 59.3  & 43.8 & 57.0 & 49.6\\ 
\Block[l]{}{\;\;\;\scriptsize{W/O Punctuation}} & 55.3 & 75.8 & 63.9 & 30.8 & 70.9 & 43.0 & 53.2 & 67.6 & 59.6  & 43.3 & 56.2 & 48.9\\
\Block[l]{}{gpt-4-0314} &  &  &  &  &  &  &  &  &  & & &  \\
\Block[l]{}{\;\;\;\scriptsize{With Punctuation}} & 57.0 & 75.9 & 65.1 & 34.0 & \textbf{73.2} & 46.4 & 54.2 & 67.7 & 60.2 & 44.2 & 57.4 & 50.0\\
\Block[l]{}{\;\;\;\scriptsize{W/O Punctuation}} & 56.4 & \textbf{76.2} & 64.8 & 31.0 & 72.0 & 43.3 & 54.2 & 69.4 & 60.9  & 45.2 & \textbf{58.2} & 51.0\\
\Block[l]{}{text-davinci-003} &  &  &  &  &  &  &  &  &  & & & \\
\Block[l]{}{\;\;\;\scriptsize{With Punctuation}} & 55.9 & 75.3 & 64.2 & 33.6 & 72.0 & 45.8 & 48.0 & 66.4 & 55.7 & 45.7 & 57.3 & 50.9\\
\Block[l]{}{\;\;\;\scriptsize{W/O Punctuation}} & 55.4 & 75.8 & 64.0 & 31.2 & 71.1 & 43.4  & 47.8 & 68.4 & 56.3 & 46.5 & 58.1 & \textbf{51.7}\\
\hline
\Block[l]{}{M2M100-1.2B} & \textbf{88.8} & 71.5 & \textbf{79.2} & \textbf{63.8} & 61.1 & \textbf{62.4} & 78.8 & \textbf{71.4} & \textbf{74.9} & 47.1 & 42.9 & 44.9 \\

\end{NiceTabular}
\caption{The results of the models on different golden tests. We report the comparative results of our best model, which is pre-trained \textit{M2M100-1.2B} fine-tuned on RUSpellRU~\cite{sorokin2016spellrueval} and MultidomainGold~\cite{martynov2023augmentation}, OpenAI models and the open source standard solutions for the Russian language. Metrics are reported in format \textbf{Prec}ision, \textbf{Rec}all, \textbf{F1}-measure from ~\cite{sorokin2016spellrueval}.}
\label{tab:resultsallgolds}
\end{table*}

\subsection{Models}

The generative models of different sizes used as pre-trained models in the experiments are the following for the Russian language:

\textbf{M2M100-1.2B}~\footnote{\url{https://huggingface.co/facebook/m2m100\_1.2B}}~\cite{DBLP:journals/corr/abs-2010-11125} M2M100 is a multilingual encoder-decoder (seq-to-seq) model primarily intended for translation tasks proposed by the Meta team. The model contains 1.2B parameters.

\textbf{M2M100-418M}~\footnote{\url{https://huggingface.co/facebook/m2m100\_418M}} is a 418M parameters model of the M2M100 models family.

\textbf{Fred-T5}~\footnote{\url{https://huggingface.co/ai-forever/FRED-T5-large}} (Full-scale Russian Enhanced Denoisers T5) is a Russian 820M parameters generative model. The model is trained on a mixture of 7 denoisers like UL2 on extensive Russian language corpus (300GB). The model is inspired by the ideas from the work~\cite{tay2023ul2} and one of the top~\footnote{\url{https://russiansuperglue.com/leaderboard/2}} generative models according to the RussianSuperGLUE benchmark~\cite{shavrina2020russiansuperglue}.

In the case of the English language, the utilization of only one pre-trained model was decided due to the considerable environmental impact caused by the training process (see section~\ref{sec:ethical} \textit{Energy Efficiency and Usage} for details).

\textbf{T5 large}~\footnote{\url{https://huggingface.co/t5-large}} is the English encoder-decoder 770M parameters model introduced by Google's AI research team~\cite{raffel2020exploring}.

\subsection{Russian experiments}

For each of the three models $M2M100-418M$, $M2M100-1.2B$, $FredT5-large$, the performance on the SC task was compared with and without pre-training, and using different training data for fine-tuning.

\textit{Pre-training.}
We use the same data and pre-training scheme for each model. We train our models in sequence-to-sequence manner with corrupted sentence as an input and correct sentence as label with a standard Cross Entropy loss. 

We pre-train $FredT5-large$ model with a total \textit{batch size} of 64, \textit{AdamW optimizer}~\cite{loshchilov2017decoupled} with an initial \textit{learning rate} of 3e-04 and \textit{linear decay} with no warm up steps and \textit{weight decay} 0.001 applied to all the parameters but those in LayerNorm~\cite{ba2016layer} and biases, and two steps to accumulate gradients for 5 \textit{epochs}. Pre-train procedure took 180 hours on eight Nvidia A100 GPUs. 

Both $M2M100-418M$ and $M2M100-1.2B$ were pre-trained with a total \textit{batch size} of 64, \textit{AdamW optimizer}~\cite{loshchilov2017decoupled} with an initial \textit{learning rate} of 5e-05, \textit{weight decay} of 0.001 applied to all the parameters but those in LayerNorm~\cite{ba2016layer} and biases, and \textit{linear decay} for learning rate without warm up steps. We also used 8 and 2 \textit{gradient accumulation steps} for $M2M100-418M$ and $M2M100-1.2B$ accordingly. $M2M100-418M$ pre-training procedure took five \textit{epochs} and 332 hours on two Nvidia A100 GPUs, and corresponding procedure for $M2M100-1.2B$ lasted for seven \textit{epochs} and 504 hours on eight  Nvidia A100 GPUs. 

\textit{Fine-tuning.}

We fine-tune pre-trained and non-pre-trained models using one of three sets: $RUSpellRU$, $MultidomainGold (MDG)$ and $RUSpellRU+MDG$. We also use the augmentation strategies for the training data presented in section~\ref{section:augmentation} and obtain additional training data to fine-tune the pre-trained models (see section~\ref{sec:datasets} Training Data for fine-tuning for details). 

We fine-tune models and take the best-performing checkpoint according to the metrics on the corresponding development set. The models’ metrics on development set is presented in the Appendix~\ref{sec:appendix:results}. We also used the development set to select the optimal hyperparameter values. We use AdamW optimizer~\cite{loshchilov2017decoupled} with $\beta_1 = 0.9$, $\beta_2 = 0.99$ and $\epsilon=1\mathrm{e}{-8}$ and a linear learning rate scheduler to fine-tune models. All hyperparameters for fine-tuning models are contained in Appendix~\ref{sec:appendix:hyperparameters}.

\textit{Model comparison.}
We compare the performance of fine-tuned models with pre-trained models in a zero-shot setting, Yandex.Speller~\footnote{\url{https://yandex.ru/dev/speller/}}, JamSpell~\footnote{\url{https://github.com/bakwc/JamSpell}}, Hunspell~\footnote{\url{https://github.com/hunspell/hunspell}}, and OpenAI~\footnote{\url{https://chat.openai.com/}} models via API (namely,\textit{ gpt-3.5-turbo-0301}, \textit{gpt4-0314}, \textit{text-davinci-003}) with different prompts (see Appendix~\ref{sec:appendix:prompts} for the prompt details) using single-domain and multi-domain test sets (see section~\ref{sec:datasets} Golden Test Sets for the details).

\subsection{English experiments}

We pre-train \textit{T5 large} model as described in ~\ref{section:pretraindata} with the following hyperparameters: % @Nikita
\textit{batch size} 64, \textit{learning rate} 3e-04 with linear decay and no warm up steps, \textit{weight decay} 0.001 applied analogously as in experiments with Russian language, 2 \textit{gradient accumulation steps}, 5 \textit{epochs}. Pre-training is done in mixed-precision with data type bfloat16~\footnote{\url{https://pytorch.org/docs/stable/generated/torch.Tensor.bfloat16.html}}. The procedure took 360 hours on eight Nvidia A100 GPUs.

We compare the performance of several models on two datasets: BEA60k and JFLEG. The models are as follows: eight NeuSpell ~\cite{jayanthi2020neuspell} models: BERT, CNN-LSTM, SC-LSTM, Nested-LSTM, SC-LSTM + BERT at input/output and SC-LSTM + ELMO at input/output. 
Additionally, we evaluate OpenAI models via API (namely, \textit{gpt-3.5-turbo-0301}, \textit{gpt4-0314}, \textit{text-davinci-003}) with different prompts: Full, Short, and Cut (see Appendix~\ref{tab:enprompts} for the details). 
Finally, we compare obtained results on the Full prompt with models from NeuSpell~\cite{jayanthi2020neuspell} and T5-large model.

\section{Evaluation}
\label{sec:evaluation}

\subsection{Metrics}

For the evaluation, we use the script from the Dialogue Shared Task~\cite{sorokin2016spellrueval}. 

As a result, the \textit{F1-measure} as the harmonic mean between \textit{Precision} and \textit{Recall} is calculated. The evaluation script reported all three metrics. 

We also evaluated models for the English language with \textit{accuracy} (correct words among all words) and \textit{correction rate} (misspelled tokens corrected), as it was proposed by~\cite{jayanthi2020neuspell}. 

\subsection{Results}

Table~\ref{tab:results_rus_generative} presents the results of experiments conducted on the Russian language. The findings indicate superior dominance of pre-trained ($+PT.$) models over the bare fine-tuning. 
Moreover, larger models generally perform better though this trend is only observed for M2M100 models. The Fred-T5 model, despite its larger size compared to the M2M100-418 model, demonstrates poorer quality on $RuspellRU$ and $MedSpellChecker$ datasets. This difference in performance may be attributed to the multilingual architecture of the M2M100 model.
In our experimental setup, we emulated errors in the pre-trained models using the $RuspellRU$ dataset. This may cause the scores of the models on this specific domain to be substantially higher than those obtained on other datasets.

Including corruption strategies (Table~\ref{tab:results_all_augmentations}) during the fine-tuning stage improves scores. This trend persists consistently across different domains. In the case of heuristic-based approach, \textit{Add} strategy celebrates most of the performance improvements. In contrast, the statistic-based approach manifests equal contribution of both strategies. 

Table~\ref{tab:resultsallgolds} demonstrates that non-generative models in the Russian language perform comparably to generative OpenAI models, but they are lightweight and more efficient. However, our best M2M100 model configuration significantly outperforms these solutions.

According to Table~\ref{tab:results_eng_golds}, the pre-trained T5 model shows comparable with OpenAI models results. We emulated the error distribution based on the BEA60K set during pre-training. However, the final evaluation of the JFLEG set is slightly better than the BEA60K.

The Tables~\ref{tab:enprompts},\ref{tab:prompts} presented in the Appendix~\ref{sec:appendix:results} demonstrate a notable gap in performance between OpenAI models for English and Russian. In English, the results indicate higher performance when punctuation is not considered. Furthermore, three models demonstrate comparable performance across all models, employing more specific prompts shows better results. However, for Russian the \textit{text-davinci-003} model with punctuation performs better. While analyzing the results, we observed that the generated outputs are sensitive to the prompts. The results contain clichés phrases, forcing additional filtering to obtain accurate results. The observed discrepancy can be attributed to the pre-trained nature of the OpenAI models primarily trained on English language data.

\section{Conclusion}
\label{sec:conclusion}
% Alena
In this paper, we have presented a novel methodology for generative SC. Our approach, which involves emulating natural spelling errors during large generative model pre-training, has shown state-of-the-art results in addressing text editing tasks. We use two augmentation techniques for text corruption to improve the results. Conducting the experiments in two languages, we have demonstrated the effectiveness of these techniques and the impact of different corruption strategies across different domains. As our research's practical impact, we proposed the library SAGE~\footnote{\url{https://github.com/ai-forever/sage/}} (that includes the data hub resource for the Russian language) for automatic SC with the proposed methods and the family of generative models. We believe our work contributes significantly to the SC field and opens routes for further exploration.

\section*{Limitations}
\label{limitations}

The proposed generative methodology of spell checking and the created models have certain limitations that should be considered:

\paragraph{Decoding strategies.} The choice of the decoding strategy affects the quality of generated texts~\cite{ippolito2019comparison}. However, our current methodology fails to comprise the entire spectrum of decoding strategies, limiting our evaluation's extent. We leave this aspect for future work.

\paragraph{Parameters.} During the pre-training and fine-tuning stages, the choice of each model's parameters is limited due to the significant computational costs associated with training and processing. Consequently, there is a potential for improved results by exploring and optimizing new parameter configurations.

\paragraph{Text Corruptions.} The heuristic approach only covers some of the augmentation methods. To address this, we plan to expand the range of hyperparameter methods for substitutions in future research. Furthermore, the different percentages of the additive noise in the data may significantly vary the result. Thus, it's a good way for future research.

\paragraph{Data collection.} A limitation of our study is the availability of different data for both the training and fine-tuning stages and the annotated data. The data used in our research may be limited to specific domains, preventing comprehensive coverage of all possible text variations. Despite these limitations, we tried to address the issue of data diversity by incorporating single-domain and multi-domain datasets in the proposed research. This approach allowed us to shed light on the diversity and variances within the data, providing valuable insights despite the inherent constraints.

\paragraph{Context.} The spell checking model's understanding and processing of word context may be limited due to the two main factors. Firstly, the model's context length is constrained (for example, T5 is limited for 512 sequence length). Secondly, the data used for the fine-tuning is limited to the text's length of the examples in the dataset, which can lead to bad performance on longer texts if the models saw only short ones. We added the domains of various text lengths to address this problem in the MultidomainGold set. Additionally, it should be mentioned that handling longer texts becomes problematic, requiring substantial computational GPU resources.

\paragraph{Languages.} The methodology employed in our study primarily focuses on investigating the applicability of our spell checking methodology within the Russian language, with an examination of its transferability to the English language. However, the generalizability of the method across diverse language families remains unclear. Thus, further research is needed to expand the datasets and evaluate the methodology's effectiveness for a wider range of languages.

\section*{Ethics Statement}
\label{sec:ethical}

In conducting our research on automatic generative SC, we recognize the importance of addressing potential ethical implications and ensuring responsible use of the developed technology. We have taken the following steps to maintain ethical standards throughout the study.

\paragraph{Crowdsourcing annotation.}
Responses of human annotators are collected and stored anonymously, eliminating personally identifiable information. The annotators are warned about potentially sensitive topics in data (e.g., politics, culture, and religion). The average annotation pay rate exceeds the hourly minimum wage in Russia twice.

\paragraph{Datasets.} We clearly state our work's aims and implications, making it open source and transparent. The data will be available under a public license. As our research involved anonymized textual data, informed consent from human participants was not required. However, we obtained permission to access publicly available datasets and ensured compliance with any applicable terms of service or usage policies.

\paragraph{Energy Efficiency and Usage.}
% CO2!!! 
Training large-scale language models consumes significant amounts of computational resources and energy, resulting in substantial carbon emissions. To minimize the ecological footprint of the research, the decision was made to limit the number of pre-trained models employed for the English language. The CO2 emission of pre-training the M2M100~\cite{fan2021beyond} and T5~\cite{raffel2020exploring} models in our experiments is computed as Equation~\ref{eq:co2}~\cite{strubell2019energy}:

\vspace{-5pt}
\begin{equation}\label{eq:co2}
    CO2 = \frac{PUE * kWh * I^{CO2}}{1000}
\end{equation}
The resulting CO2 emissions are listed below:
\begin{enumerate}
    \item \textit{M2M100-1.2B} = 87.09 kg;
    \item \textit{M2M100-418M} = 57.37 kg;
    \item \textit{T5-large} = 62.21 kg;
    \item \textit{FredT5-large} = 31.11 kg;
\end{enumerate}

The power usage effectiveness ($PUE$) of our data centers is not more than $1.3$.
Despite the costs, spelling models can be efficiently adapted to the user needs and bringing down potential budget costs in the scope of modern applications.
Model compression techniques, e.g., pruning and distillation, can further reduce the model inference cost and footprint.

\paragraph{Biases.}
The datasets we collected include large segments representing the Internet domain, and therefore, they may contain various stereotypes and biases, as well as the pre-train models. The scope of risks associated with the misuse of generative language models is widely discussed in the community~\cite[]{weidinger2021ethical,bommasani2021opportunities}. We acknowledge the potential for biases to emerge in both the training data and the model's predictions. Proper evaluation is still needed to explore possible model vulnerabilities in terms of generalizing on the new data and specific new data. 

\paragraph{Possible Misuse.}
We understand that the results of our work can be used maliciously, e.g., to write inappropriate and toxic texts. We believe that our research should not be involved in creating content that affects the individual or communal well-being in any way, including legislative application or censorship; mis- and disinformation; infringement of the rights of access to information.

We propose a novel methodology applied potentially to any language and a valuable resource for the Russian language in particular. We anticipate that our work may contribute to improved written communication, but we also recognize the need for ongoing ethical evaluation to address emerging challenges.

\section*{Acknowledgements}
The authors sincerely thank Alexey Sorokin for providing us with the evaluation script from the Dialogue Shared task.
The authors would also like to extend their appreciation to the teams of the authors of the datasets we took for the training and testing parts. We thank DmitryPogrebnoy, the author of anamnesis medical data validated and included in our MedSpellChecker set.
The authors are grateful for Ibragim Badertdinov's ideas of heuristic-based corrupted method in the texts.
The authors would like to thank Denis Kulagin and his "kartaslov"~\footnote{\url{https://kartaslov.ru/}} git-project for the data and statistics on typos.
The authors are deeply grateful for the valuable contributions of everyone mentioned above. Their efforts played a crucial role in completing this research.

\bibliography{anthology,custom}
\bibliographystyle{acl_natbib}

\newpage
\clearpage
\appendix

\section{Appendix}
\label{sec:appendix}

\subsection{Data}
\label{sec:appendix:data}
The information of the collected data for the train set and expansion of the gold sets are presented in Tables~\ref{tab:traindata_statistics} and~\ref{tab:testdata_statistics}. 

\begingroup
\tabcolsep=0.175cm
\begin{table}[htbp!]
\footnotesize
\centering
\begin{tabular*}{\columnwidth}{lcccc}
\hline
\textbf{Datasets} & \textbf{1S-A} & \textbf{2S-A} & \textbf{Size}
& \textbf{Length}\\
\hline
Web (Aranea) & + & + & 756 & 133.8 \\
Literature  & + & + & 260 & 194.3 \\
News  & + & + & 245 & 278.7 \\
Social media  & + & + & 200 & 149.6 \\
Strategic Doc  & + & + & 250 & 182.9 \\ \hline
Reviews  & + & + & 586 & 678.9 \\
OpenSubtitles  & + & + & 1810 & 44.2 \\ \hline
RUSpellRU & - & - & 2008 & 87 \\
GitHubTypoCorpusRu & - & + & 868 & 156 \\ \hline
MedSpellChecker & + & + & 1054 & 135 \\ \hline
BEA60k & - & - & 63044 & 79.1 \\
JFLEG & - & - & 1601 & 109
\\ \hline
\end{tabular*}
\caption{The test golden sets statistics. The sizes of the test sets parts in the number of examples (mostly sentences). $1S-A$ represents if the dataset was validated on the first annotation step. $2S-A$ represents if the dataset was validated on the second annotation step. $Length$ is the average number of symbols in one dataset's example.}
\label{tab:testdata_statistics}
\end{table}
\endgroup

\begingroup
\tabcolsep=0.195cm
\begin{table}[htbp!]
\centering
\begin{tabular*}{\columnwidth}{lcccc}
\hline
\textbf{Datasets} & \textbf{1S-A} & \textbf{2S-A} & \textbf{Size} & \textbf{Length}\\
\hline
Web (Aranea)  & + & + & 386 & 108.4 \\
News  & + & + & 361 & 268.1\\
Social media  & + & + & 430 & 163.9 \\
OpenSubtitles  & + & + & 1810 & 45.3 \\ \hline
Reviews  & + & + & 584 & 689.1 \\
RUSpellRU & - & - & 2000 & 77.9 \\  \hline
\end{tabular*}
\caption{The train sets statistics. The sizes of the train sets parts in the number of examples (primarily sentences). $1S-A$ represents if the dataset was validated on the first annotation step. $2S-A$ represents if the dataset was validated on the second annotation step. $Length$ is the average number of symbols in one dataset's example.}
\label{tab:traindata_statistics}
\end{table}
\endgroup

\begin{table*}[htbp!]  % \label{table_results}
\centering
\footnotesize
\begin{NiceTabular}{{c}|{c}{c}{c}|{c}{c}{c}|{c}{c}{c}|*{3}{p{0.7cm}}}%[hvlines]

\hline
\Block{2-1}{\textbf{Model}} & \Block{1-3}{\textbf{RUSpellRU}} & & &  \Block{1-3}{\textbf{MultidomainGold}} & & & \Block{1-3}{\textbf{MedSpellChecker}} & & & \Block{1-3}{\textbf{GitHubTypoCorpusRu}} \\ 
& Prec. & Rec. & F1 & Prec. & Rec. & F1 & Prec. & Rec. & F1 & Prec. & Rec. & F1 \\
\hline
\Block[l]{}{\textbf{M2M100-1.2B}} & & & & & & & & & & & & \\
%\multicolumn{1}{|r|}{Item3}
\Block[l]{}{\;\;\scriptsize{Best-of-FT/PT.}} & \textbf{88.8} & 72.5 & \textbf{79.2} & \textbf{63.8} & 61.1 & 62.4 & \textbf{78.8} & 73.9 & 75.1 & 47.9 & 43.3 & 45.5\\
\Block[l]{}{\;\;\scriptsize{\underline{Augmentex (Add)}}} & & & & & & & & & & & & \\
\Block[l]{}{\;\;\;\;\scriptsize{RUSpellRU}} 
& 70.6 & 74.0 & 72.3
& 46.7 & 59.0 & 52.1
& 48.5 & 63.2 & 54.9
& 40.9 & 44.7 & 42.7 \\
\Block[l]{}{\;\;\;\;\scriptsize{MultidomainGold}} 
& 73.7 & 67.4 & 70.4
& 58.1 & 62.0 & 60.0
& 69.4 & 74.2 & 71.7
& 47.8 & 47.1 & 47.5\\
\Block[l]{}{\;\;\;\;\scriptsize{RUSpellRU+MDG}} 
& 75.9 & 75.7 & 75.8
& 57.4 & \textbf{64.8} & 60.9
& 63.3 & 72.9 & 67.8
& 48.0 & \textbf{48.1} & \textbf{48.1} \\
\Block[l]{}{\;\;\scriptsize{\underline{Augmentex (Concat.)}}} & & & & & & & & & & & & \\
\Block[l]{}{\;\;\;\;\scriptsize{RUSpellRU}} 
& 72.8 & 75.4 & 74.0
& 48.4 & 60.3 & 53.7
& 49.9 & 63.7 & 56.0
& 41.5 & 45.7 & 43.5 \\
\Block[l]{}{\;\;\;\;\scriptsize{MultidomainGold}} 
& 76.7 & 68.6 & 72.4
& 60.8 & 63.0 & 61.9
& 69.4 & 71.9 & 70.6
& 48.4 & 45.5 & 46.9 \\
\Block[l]{}{\;\;\;\;\scriptsize{RUSpellRU+MDG}} 
& 79.3 & \textbf{76.5} & 77.9
& 59.6 & 63.6 & 61.5
& 68.5 & 72.1 & 70.2
& 48.4 & 47.0 & 47.7 \\
\Block[l]{}{\;\;\scriptsize{\underline{SBSC (Add)}}} & & & & & & & & & & & & \\
\Block[l]{}{\;\;\;\;\scriptsize{RUSpellRU}} 
& 79.0 & 74.2 & 76.6
& 52.0 & 59.2 & 55.4
& 53.0 & 58.8 & 55.8
& 37.7 & 42.7 & 40.0 \\
\Block[l]{}{\;\;\;\;\scriptsize{MultidomainGold}} 
& 86.0 & 60.6 & 71.1
& 63.7 & 63.1 & \textbf{63.4}
& 77.4 & \textbf{75.2} & \textbf{76.3}
& 47.5 & 41.4 & 44.2\\
\Block[l]{}{\;\;\;\;\scriptsize{RUSpellRU+MDG}} 
& 84.0 & 74.7 & 79.1
& 61.2 & 64.4 & 62.8
& 73.3 & 72.4 & 72.8
& 47.2 & 43.3 & 45.2\\
\Block[l]{}{\;\;\scriptsize{\underline{SBSC (Concat.)}}} & & & & & & & & & & & & \\
\Block[l]{}{\;\;\;\;\scriptsize{RUSpellRU}} 
& 83.3 & 72.3 & 77.4
& 54.0 & 59.4 & 56.6
& 64.7 & 56.3 & 60.2
& 41.7 & 41.8 & 41.7 \\
\Block[l]{}{\;\;\;\;\scriptsize{MultidomainGold}} 
& 82.8 & 66.3 & 73.6
& 63.5 & 63.3 & \textbf{63.4}
& 74.3 & 71.6 & 72.9
& \textbf{48.6} & 44.5 & 46.5\\
\Block[l]{}{\;\;\;\;\scriptsize{RUSpellRU+MDG}} 
& 85.9 & 72.5 & 78.6
& 62.5 & 63.3 & 62.9
& 73.9 & 68.0 & 70.8
& 47.7 & 43.1 & 45.3\\
\hline
\Block[l]{}{\textbf{M2M100-418M}} & & & & & & & & & & & & \\
%\multicolumn{1}{|r|}{Item3}
\Block[l]{}{\;\;\scriptsize{Best-of-FT/PT.}} & \textbf{87.6} & 64.4 & \textbf{74.2} & \textbf{60.3} & 56.6 & \textbf{58.4} & \textbf{73.5} & 66.0 & \textbf{69.5} & 42.8 & 42.6 & 40.2\\
\Block[l]{}{\;\;\scriptsize{\underline{Augmentex (Add)}}} & & & & & & & & & & & & \\
\Block[l]{}{\;\;\;\;\scriptsize{RUSpellRU}} 
& 60.1 & 71.2 & 65.1
& 35.2 & 64.1 & 45.5
& 24.0 & 58.6 & 34.1
& 28.3 & 45.8 & 35.0 \\
\Block[l]{}{\;\;\;\;\scriptsize{MultidomainGold}} 
& 61.2 & 66.6 & 63.8
& 49.0 & 61.1 & 54.4
& 48.4 & \textbf{70.1} & 57.3
& 41.0 & 46.3 & 43.5 \\
\Block[l]{}{\;\;\;\;\scriptsize{RUSpellRU+MDG}} 
& 63.1 & 70.8 & 66.7
& 47.4 & 60.4 & 53.1
& 48.6 & 68.5 & 56.8
& 41.3 & \textbf{47.0} & \textbf{44.0} \\
\Block[l]{}{\;\;\scriptsize{\underline{Augmentex (Concat.)}}} & & & & & & & & & & & & \\
\Block[l]{}{\;\;\;\;\scriptsize{RUSpellRU}} 
& 65.5 & \textbf{71.3} & 68.3
& 38.0 & \textbf{64.5} & 47.8
& 28.1 & 60.1 & 38.3
& 29.8 & 44.4 & 35.7 \\
\Block[l]{}{\;\;\;\;\scriptsize{MultidomainGold}} 
& 68.7 & 64.9 & 66.7
& 54.2 & 60.2 & 57.0
& 58.1 & 66.8 & 62.1
& \textbf{42.9} & 43.3 & 43.1 \\
\Block[l]{}{\;\;\;\;\scriptsize{RUSpellRU+MDG}} 
& 73.1 & 70.2 & 71.7
& 55.0 & 60.3 & 57.5
& 56.1 & 68.3 & 61.6
& \textbf{42.9} & 42.8 & 42.8 \\
\Block[l]{}{\;\;\scriptsize{\underline{SBSC (Add)}}} & & & & & & & & & & & & \\
\Block[l]{}{\;\;\;\;\scriptsize{RUSpellRU}} 
& 75.7 & 67.5 & 71.4
& 43.2 & 59.9 & 50.2
& 36.9 & 56.0 & 44.5
& 31.8 & 41.5 & 36.0\\
\Block[l]{}{\;\;\;\;\scriptsize{MultidomainGold}} 
& 75.5 & 61.2 & 67.6
& 55.1 & 57.9 & 56.5
& 65.0 & 67.0 & 66.0
& 42.4 & 42.0 & 42.2\\
\Block[l]{}{\;\;\;\;\scriptsize{RUSpellRU+MDG}} 
& 78.2 & 67.7 & 72.6
& 56.4 & 59.9 & 58.1
& 64.5 & 67.3 & 65.8
& 42.1 & 40.3 & 41.2\\
\Block[l]{}{\;\;\scriptsize{\underline{SBSC (Concat.)}}} & & & & & & & & & & & & \\
\Block[l]{}{\;\;\;\;\scriptsize{RUSpellRU}} 
& 79.5 & 65.8 & 72.0
& 46.4 & 58.5 & 51.8
& 43.8 & 53.2 & 48.0
& 31.4 & 37.2 & 34.0\\
\Block[l]{}{\;\;\;\;\scriptsize{MultidomainGold}} 
& 75.2 & 56.5 & 64.5
& 55.9 & 54.0 & 55.0
& 64.9 & 61.4 & 63.1
& 42.1 & 41.2 & 41.6\\
\Block[l]{}{\;\;\;\;\scriptsize{RUSpellRU+MDG}} 
& 83.6 & 65.6 & 73.5
& 58.7 & 55.4 & 57.0
& 66.8 & 64.5 & 65.6
& 42.5 & 39.0 & 40.7\\
\hline
\Block[l]{}{\textbf{FredT5-large}} & & & & & & & & & & & & \\
%\multicolumn{1}{|r|}{Item3}
\Block[l]{}{\;\;\scriptsize{Best-of-FT/PT.}} & 74.5 & 73.4 & 73.9 & 61.7 & 63.1 & \textbf{61.1} & 43.2 & 60.4 & 47.7 & \textbf{69.3} & 45.4 & 54.3\\
\Block[l]{}{\;\;\scriptsize{\underline{Augmentex (Add)}}} & & & & & & & & & & & & \\
\Block[l]{}{\;\;\;\;\scriptsize{RUSpellRU}} 
& 51.9 & 74.6 & 61.2
& 25.0 & 57.5 & 34.9
& 12.3 & 51.4 & 19.8
& 25.4 & 43.7 & 32.2\\
\Block[l]{}{\;\;\;\;\scriptsize{MultidomainGold}} 
& 67.4 & 67.4 & 67.4
& 55.8 & 62.6 & 59.0
& 36.6 & 60.1 & 45.5
& 61.4 & 47.7 & 53.7 \\
\Block[l]{}{\;\;\;\;\scriptsize{RUSpellRU+MDG}} 
& 72.0 & \textbf{77.9} & \textbf{74.8}
& 51.9 & \textbf{66.6} & 58.3
& 36.5 & 61.4 & 45.8
& 56.7 & \textbf{49.3} & 52.7 \\
\Block[l]{}{\;\;\scriptsize{\underline{Augmentex (Concat.)}}} & & & & & & & & & & & & \\
\Block[l]{}{\;\;\;\;\scriptsize{RUSpellRU}} 
& 53.3 & 75.6 & 62.5
& 26.6 & 59.2 & 36.7
& 12.5 & 51.7 & 20.1
& 26.1 & 44.0 & 32.8 \\
\Block[l]{}{\;\;\;\;\scriptsize{MultidomainGold}} 
& 66.1 & 67.2 & 66.7
& 55.5 & 65.7 & 60.2
& 36.6 & 64.5 & 46.7
& 64.4 & 47.9 & \textbf{54.9} \\
\Block[l]{}{\;\;\;\;\scriptsize{RUSpellRU+MDG}} 
& 71.1 & 75.0 & 73.0
& 51.1 & 62.6 & 56.3
& 34.9 & 58.1 & 43.6
& 60.3 & 48.0 & 53.5 \\
\Block[l]{}{\;\;\scriptsize{\underline{SBSC (Add)}}} & & & & & & & & & & & & \\
\Block[l]{}{\;\;\;\;\scriptsize{RUSpellRU}} 
& 54.5 & 73.4 & 62.5
& 27.1 & 57.0 & 36.8
& 13.0 & 51.2 & 20.8
& 25.9 & 41.3 & 31.8 \\
\Block[l]{}{\;\;\;\;\scriptsize{MultidomainGold}} 
& 73.5 & 59.3 & 65.7
& 61.5 & 60.5 & 61.0
& \textbf{47.6} & 57.0 & 51.9
& 66.8 & 44.6 & 53.5\\
\Block[l]{}{\;\;\;\;\scriptsize{RUSpellRU+MDG}} 
& \textbf{77.4} & 71.4 & 74.3
& 57.8 & 61.5 & 59.6
& 41.6 & 57.5 & 48.3
& 60.1 & 46.0 & 52.1\\
\Block[l]{}{\;\;\scriptsize{\underline{SBSC (Concat.)}}} & & & & & & & & & & & & \\
\Block[l]{}{\;\;\;\;\scriptsize{RUSpellRU}} 
& 55.0 & 69.8 & 61.5
& 26.0 & 53.5 & 35.0
& 12.8 & 47.1 & 20.1
& 27.4 & 41.3 & 32.9 \\
\Block[l]{}{\;\;\;\;\scriptsize{MultidomainGold}} 
& 64.8 & 63.1 & 64.0
& 59.0 & 62.7 & 60.8
& 38.6 & \textbf{65.2} & 48.5
& 62.6 & 46.0 & 53.0 \\
\Block[l]{}{\;\;\;\;\scriptsize{RUSpellRU+MDG}} 
& 72.4 & 74.6 & 73.5
& \textbf{61.7} & 60.2 & 61.0
& 42.7 & 58.6 & \textbf{49.4}
& 65.4 & 46.2 & 54.1 \\ \hline
\end{NiceTabular}

\caption{Pre-trained models' performance on test datasets for the Russian language after fine-tuning on augmented datasets. \textit{Augmentex} and \textit{SBSC} represent different methods of augmentation described in~\ref{section:augmentation}. \textit{Add} and \textit{Concat.} represent different strategies of augmentation described in~\ref{sec:datasets} in the section Training Data for fine-tuning. Metrics reported in format \textbf{Prec}ision, \textbf{Rec}all, \textbf{F1} from~\cite{sorokin2016spellrueval}.}
\label{tab:results_all_augmentations}
\end{table*}

\subsection{Annotation}
\label{sec:appendix:annotation}

For the extension of the gold test set and the MultidomainGold train part, we use the two-stage annotation setups via a crowd-sourcing platform Toloka\footnote{\url{https://toloka.ai/tolokers}}~\cite{toloka} similarly to the work~\cite{martynov2023augmentation}:
\begin{enumerate}
    \item \textbf{Data gathering stage}: the texts with possible mistakes are provided, and the annotators are asked to write the sentence correctly; 
    \item \textbf{Validation stage}: the pair of sentences (source and its corresponding correction from the previous stage) are provided, and the annotators are asked to check if the correction is right.
\end{enumerate}

The annotation costs and the details for the created sets in the current work are presented in Table~\ref{tab:toloka_details}. 

\begin{table}[htbp!]
\centering
\begin{tabular}{lcccc}
\hline
    \textbf{Params}  & \textbf{S1.Tr} & \textbf{S2.Tr}  & \textbf{S1.Te}  & \textbf{S2.Te} \\
    \hline
    \textbf{IAA}    & 82.06  & 85.20 & 82.34  & 91.78    \\
    \textbf{Total}    & 720\$  & 451\$   & 732\$  &  947\$   \\
    \textbf{Overlap}    & 3  & 3  & 3  &  3  \\
    \textbf{$N_T$}     & 7  & 7  &  8 & 8   \\
    \textbf{$N_{page}$}   & 4  & 5 & 4  & 5   \\
    \textbf{$N_C$}   & 50  & 46 & 50  & 46  \\
    \textbf{$N_U$}   & 12  & 10 & 10 &  9  \\
    \textbf{ART}   & 102  & 71 & 95 &  60   \\
    \hline
\end{tabular}

    \caption{Details on the data collection projects for the Golden Test sets and the Train MultidomainGold for both parts of the annotation pipeline ($S1.Tr$ is first annotation stage of train set; $S2.Te$ is second annotation step of the testset respectively). \textbf{IAA} refers to the average IAA confidence scores, \%. IAA of the first step is calculated as the expected value of annotators’ support of the most popular correction over all labeled texts. IAA of second step is calculated as an average value of confidence scores over all labeled texts. \textbf{Total} is the total cost of the annotation project. \textbf{Overlap} is the number of votes per example. $N_T$ is the number of training tasks. $N_{page}$ denotes the number of examples per page. $N_C$ is the number of control examples. $N_U$ is the number of users who annotated the tasks. \textbf{ART} means the average response time in seconds.}
    \label{tab:toloka_details}
\end{table}

\begin{table}[htbp!]
\centering
\begin{tabular}{lccc}
\hline
\textbf{Model} & \textbf{Speed} & \textbf{Size} & \textbf{Params}\\
\hline
M2M100-1.2B & 175.73 & 4.96 & 1.2B \\
M2M100-418  & 326.16 & 1.94 & 418M \\
Fred-T5-large & 177.12 & 3.28 & 820M \\
T5-large  & 190.96 & 2.95 & 770M \\ \hline
\end{tabular}
\caption{The Models' statistics. $Speed$ is the speed of the model on inference on a single Nvidia A100 in symbols per second. $Params$ represents the number of the models' parameters. $Size$ is the size of the models' checkpoint weights in GB.}
\label{tab:model_statistics}
\end{table}

\subsection{Augmentation strategies details}
\label{sec:appendix:augmentations}
In the diverse array of settings available within Augmentex, customization options include the percentage of phrase changes, the maximum and minimum number of errors, and the proportion of phrases eligible for modifications. Among its various augmentation strategies, we choose the word-level approach (replacing the symbols with a probability of their mistaken use) and the sentence-level approach (substituting words with frequent incorrect alternatives). We configured the first setup with the parameters: aug\_rate=0.1, min\_aug=1, max\_aug=3, mult\_num=5, action="orfo" and aug\_prob=0.7, and the second: aug\_rate=0.6, min\_aug=1, max\_aug=5, action="replace" and aug\_prob=0.7.

\subsection{Experiments evaluation results}
\label{sec:appendix:results}
The evaluation of all the experiments discussed in the section~\ref{sec:experiments} are presented in the Tables~\ref{tab:enprompts},~\ref{tab:prompts},~\ref{tab:results_all_augmentations},~\ref{tab:results_eng_golds}.
The evaluation on development sets during the training is presented in the Table~\ref{tab:results_dev_set}.

\begin{table*}[htbp!]
\centering
\footnotesize
\begin{NiceTabular}{{c}|{c}{c}{c}|{c}{c}{c}|{c}{c}{c}}
\hline
\Block{2-1}{} & \Block{1-3}{\textbf{M2M100-1.2B}} & & &  \Block{1-3}{\textbf{M2M100-418M}} & & & \Block{1-3}{\textbf{FredT5-large}} \\ 
& Prec. & Rec. & F1 & Prec. & Rec. & F1 & Prec. & Rec. & F1 \\
\hline
\Block[l]{}{\textbf{Fine-tuning}} & & & & & & & & & \\
%\multicolumn{1}{|r|}{Item3}
\Block[l]{}{\;\;\scriptsize{\underline{without Pre-training}}} & & & & & & & & & \\
\Block[l]{}{\;\;\;\;\scriptsize{RUSpellRU}} 
& 70.8 & 53.1 & 60.6 
& 70.5 & 50.0 & 58.5 
& 35.6 & 58.2 & 44.2 \\
\Block[l]{}{\;\;\;\;\scriptsize{MultidomainGold}} 
& 40.0 & 41.2 & 40.6 
& 34.7 & 40.5 & 37.4 
& 51.3 & 52.8 & 52.1 \\
\Block[l]{}{\;\;\;\;\scriptsize{RUSpellRU+MDG}} 
& 51.9 & 45.6 & 48.5 
& 46.7 & 45.8 & 46.3 
& 48.5 & 57.0 & 52.4 \\
\Block[l]{}{\;\;\scriptsize{\underline{with Pre-training}}} & & & & & & & & & \\
\Block[l]{}{\;\;\;\;\scriptsize{RUSpellRU}}
& \textbf{88.5} & \textbf{82.7} & \textbf{85.5} & \textbf{80.2} & \textbf{72.5} & \textbf{76.1} & 46.7 & \textbf{80.1} & 59.0 \\
\Block[l]{}{\;\;\;\;\scriptsize{MultidomainGold}} 
& 60.2 & 67.8 & 63.8 & 52.5 & 59.8 & 55.9 & 62.1 & 69.8 & 65.7 \\
\Block[l]{}{\;\;\;\;\scriptsize{RUSpellRU+MDG}} 
& 72.2 & 73.6 & 72.9 & 64.2 & 64.2 & 64.2 & \textbf{62.9} & 75.7 & \textbf{68.7} \\
\hline
\Block[l]{}{\textbf{Augmentations}} & & & & & & & & & \\
%\multicolumn{1}{|r|}{Item3}
\Block[l]{}{\;\;\scriptsize{\underline{Augmentex (Add)}}} & & & & & & & & \\
\Block[l]{}{\;\;\;\;\scriptsize{RUSpellRU}} 
& 82.7 & 82.7 & 82.7 & 66.1 & 76.5 & 70.9 & 44.7 & 78.1 & 56.9 \\
\Block[l]{}{\;\;\;\;\scriptsize{MultidomainGold}} 
& 58.3 & 68.8 & 63.1 & 44.2 & 63.3 & 52.1 & 56.7 & 70.1 & 62.7 \\
\Block[l]{}{\;\;\;\;\scriptsize{RUSpellRU+MDG}} 
& 67.5 & 78.5 & 72.6 & 53.1 & 71.3 & 60.9 & 56.6 & 77.3 & 65.4 \\
\Block[l]{}{\;\;\scriptsize{\underline{Augmentex (Concat.)}}} & & & & & & & & \\
\Block[l]{}{\;\;\;\;\scriptsize{RUSpellRU}} 
& 82.7 & 82.7 & 82.7 & 71.2 & 78.1 & 74.5 & 46.4 & \textbf{81.6} & 59.2 \\
\Block[l]{}{\;\;\;\;\scriptsize{MultidomainGold}} 
& 58.8 & 69.8 & 63.8 & 48.3 & 61.8 & 54.2 & 54.1 & 73.1 & 62.2 \\
\Block[l]{}{\;\;\;\;\scriptsize{RUSpellRU+MDG}} 
& 68.7 & 76.9 & 72.6 & 56.7 & 68.0 & 61.9 & 56.7 & 76.3 & 65.0 \\
\Block[l]{}{\;\;\scriptsize{\underline{SBSC (Add)}}} & & & & & & & & & \\
\Block[l]{}{\;\;\;\;\scriptsize{RUSpellRU}} 
& \textbf{88.6} & 83.2 & \textbf{85.8} & 77.5 & \textbf{79.1} & \textbf{78.3} & 46.3 & 78.6 & 58.2 \\
\Block[l]{}{\;\;\;\;\scriptsize{MultidomainGold}} 
& 57.5 & 68.8 & 62.6 & 50.3 & 63.1 & 56.0 & 63.5 & 72.8 & 67.8 \\
\Block[l]{}{\;\;\;\;\scriptsize{RUSpellRU+MDG}} 
& 69.8 & 76.9 & 73.2 & 59.4 & 69.8 & 64.2 & 63.3 & 76.7 & 69.3 \\
\Block[l]{}{\;\;\scriptsize{\underline{SBSC (Concat.)}}} & & & & & & & & & \\
\Block[l]{}{\;\;\;\;\scriptsize{RUSpellRU}} 
& 86.8 & \textbf{84.2} & 85.5 & \textbf{79.7} & 76.0 & 77.8 & 45.2 & 78.6 & 57.4 \\
\Block[l]{}{\;\;\;\;\scriptsize{MultidomainGold}} 
& 59.8 & 69.1 & 64.1 & 51.1 & 60.5 & 55.4 & 61.2 & 71.7 & 66.1 \\
\Block[l]{}{\;\;\;\;\scriptsize{RUSpellRU+MDG}} 
& 68.4 & 76.5 & 72.2 & 62.5 & 65.8 & 64.1 & \textbf{66.0} & 76.7 & \textbf{71.0} \\ \hline
\end{NiceTabular}

\caption{The evaluation of models' configurations with fine-tuning and the augmentations on dev sets. Metrics are reported in format \textbf{Prec}ision, \textbf{Rec}all, \textbf{F1}-measure from ~\cite{sorokin2016spellrueval}}
\label{tab:results_dev_set}
\end{table*}

\begin{figure}[htbp!]
\centering
\includegraphics[width=\columnwidth]{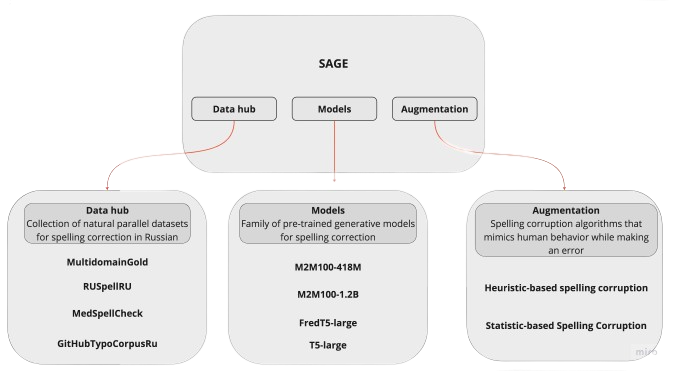}
\caption{The architecture overview of the SAGE library.}
\label{fig:architecture overview}
\end{figure}

\subsection{SAGE library}
\label{sec:appendix:library}
As the practical result of the introduced methodology, we present SAGE~\footnote{\url{https://github.com/ai-forever/sage/}} (Spell checking via Augmentation and  Generative distribution Emulation). The library consists of three parts: data hub, augmentation strategies, and the family of the models. The architecture is presented on a schema~\ref{fig:architecture overview}. The data hub includes the whole collection of natural parallel datasets for SC in Russian that was created within the frame of our research. The family of pre-trained generative models for SC involves all the best models trained during the current research for the Russian and English languages. The models are assessed with the inference code from the HuggingFace library~\footnote{\url{https://github.com/huggingface/transformers}} and the evaluation script. The last part is the augmentation methods included in SAGE. The statistic-based approach is presented for emulating the user’s parallel corpus distribution and provides the emulation algorithm on new data. The heuristic-based approach is presented for producing the noise via different strategies on a word and sentence level in the non-labeled text data.

\begin{table*}[htbp!]  % \label{table_results}
\centering
\footnotesize
\begin{NiceTabular}{{c}|{c}{c}{c}|{c}{c}|{c}{c}{c}|{c}{c}}%[hvlines]
\hline
\Block{2-1}{\textbf{Model}} & \Block{1-5}{\textbf{BEA60K}} & & & & & \Block{1-5}{\textbf{JFLEG}} \\ 
& Prec. & Rec. & F1 & Acc. &  Cor. rate & Prec. & Rec. & F1 & Acc. &  Cor. rate \\
\hline
\Block[l]{}{BERT} & 65.8 & 79.6 & 72.0 & \textbf{0.98} & 0.79 & 78.5 & 85.4 & 81.8 & \textbf{0.98} & \textbf{0.85} \\
\Block[l]{}{CNN-LSTM} & 59.7 & 76.0 & 66.8 & 0.96 & 0.76 & 76.8 & 81.1 & 78.9 & \textbf{0.98} & 0.80 \\
\Block[l]{}{SC-LSTM} & 61.7 & 77.1 & 68.6 & 0.96 & 0.77 & 77.6 & 82.1 & 79.8 & \textbf{0.98} & 0.82 \\
\Block[l]{}{Nested-LSTM} & 63.1 & 77.7 & 69.7 & 0.96 & 0.77 & 78.7 & 82.7 & 80.6 & \textbf{0.98} & 0.82 \\
\hline
\Block[l]{}{SC-LSTM}  \\
%\multicolumn{1}{|r|}{Item3}
\Block[l]{}{\;\;\;\scriptsize{+BERT (at input)}} 
& 66.2 & 77.5 & 71.4 & \textbf{0.98} & 0.77 & 78.1 & 83.0 & 80.5 & \textbf{0.98} & 0.83\\
\Block[l]{}{\;\;\;\scriptsize{+BERT (at output)}} 
& 64.1 & 76.7 & 69.8 & 0.97 & 0.76 & 78.3 & 83.2 & 80.6 & \textbf{0.98} & 0.83\\
\Block[l]{}{\;\;\;\scriptsize{+ELMO (at input)}} 
& 62.3 & 80.4 & 72.0 & 0.96 &\textbf{0.80} & 80.6 & 86.1 & 83.3 & 0.98 & \textbf{0.85}\\
\Block[l]{}{\;\;\;\scriptsize{+ELMO (at input)}} 
& 60.4 & 76.5 & 67.5 & 0.96 & 0.77 & 77.7 & 82.5 & 80.0 & \textbf{0.98} & 0.82\\
\hline
\Block[l]{}{gpt-3.5-turbo-0301} &  &  &  &  &  &  &  &  &  &  \\
\Block[l]{}{\;\;\;\scriptsize{W/O Punctuation}} & \textbf{66.9} & 84.1 & 74.5 & 0.84 & 0.77 & 77.8 & 88.6 & 82.9 & 0.87 & 0.78\\
\Block[l]{}{\;\;\;\scriptsize{With Punctuation}} & 57.1 & 83.5 & 67.8 & 0.36 & 0.34 & 73.3 & \textbf{87.9} & 80.0 & 0.34 & 0.32 \\
\Block[l]{}{gpt-4-0314} &  &  &  &  &  &  &  &  &  &  \\
\Block[l]{}{\;\;\;\scriptsize{W/O Punctuation}} & 68.6 & \textbf{85.2} & \textbf{76.0} & 0.84 & 0.77 & 77.9 & 88.3 & 82.8 & 0.86 & 0.77 \\
\Block[l]{}{\;\;\;\scriptsize{With Punctuation}} & 58.4 & 84.5 & 69.1 & 0.36 & 0.35 & 73.5 & 87.7 & 80.0 & 0.35 & 0.32  \\
\Block[l]{}{text-davinci-003} &  &  &  &  &  &  &  &  &  &  \\
\Block[l]{}{\;\;\;\scriptsize{W/O Punctuation}} & 67.8 & 83.9 & 75.0 & 0.83 & 0.76 & 76.8 & 88.5 & 82.2 & 0.87 & 0.78 \\
\Block[l]{}{\;\;\;\scriptsize{With Punctuation}} & 57.6 & 83.3 & 68.1 & 0.35 & 0.34 & 72.7 & \textbf{87.9} & 79.6 & 0.34 & 0.32 \\
\hline
\Block[l]{}{T5-large (+PT.)} & 66.5 & 83.1 & 73.9 & 0.83 & 0.71 & \textbf{83.4} & 84.3 & \textbf{83.8} & 0.74 & 0.69 \\ \hline
\end{NiceTabular}

\caption{The models' performance for the English language on BEA60K and JFLEG datasets. We report the comparative results of our best model, OpenAI models and the open source standard solutions for the English language. Metrics are reported in \textbf{Prec}ision / \textbf{Rec}all / \textbf{F1}-measure and \textbf{Acc}uracy / \textbf{Cor}rection rate formats from ~\cite{sorokin2016spellrueval} and \cite{jayanthi2020neuspell} respectively.}
\label{tab:results_eng_golds}
\end{table*}

\subsection{OpenAI models prompts experiments}
\label{sec:appendix:prompts}
We conduct experiments~\ref{tab:enprompts},~\ref{tab:prompts} varying different prompts OpenAI models to evaluate their performance on Golden test sets in Russian and English. For both English and Russian sets, we try three types of prompts:
1) \textbf{Cut prompt} for Russian: "Perepishi tekst bez orfograficheskih, grammaticheskih oshibok i opechatok, sohranjaja ishodnyj stil' teksta, punktuaciju, ne raskryvaja abbreviatur i ne izmenjaja korrektnyj tekst:"; for English: "Correct spelling and grammar in the following text:".
2) \textbf{Short prompt} for Russian: "Perepishi tekst bez orfograficheskih, grammaticheskih oshibok i opechatok, sohranjaja ishodnyj stil' teksta, punktuaciju, ne raskryvaja abbreviatur i ne izmenjaja korrektnyj tekst:"; for English: "Correct spelling and grammar in the following text: . Do not provide any interpretation of your answer.".
3) \textbf{Full Prompt} for Russian: "Perepishi tekst bez orfograficheskih, grammaticheskih oshibok i opechatok, sohranjaja ishodnyj stil' teksta, punktuaciju, ne raskryvaja abbreviatur, ne izmenjaja korrektnyj tekst. Napishi tol'ko pravil'nyj otvet bez dopolnitel'nyh ob"jasnenij."; for English: "Rewrite text without spelling errors, grammatical errors and typos, preserve the original text style, punctuation, do not open abbreviations and do not change the correct text. Do not provide any interpretation of your answer.".
% примеры промптов! Катя + Алена 

\begingroup
\tabcolsep=0.16cm 
\begin{table*}[htbp!]
\centering
\footnotesize
\begin{NiceTabular*}{\textwidth}{{c}|*{3}{c}|*{3}{c}|*{3}{c}|*{3}{c}|*{3}{c}|*{3}{c}}%
\hline
\Block{3-1}{\textbf{Prompt}} & & & \Block{1-2}{\textbf{gpt-3.5-turbo-0301}} &  & &  & &  & \Block{1-2}{\textbf{gpt-4-0314}} & &  & & & & \Block{1-2}{\textbf{text-davinci-003}} \\ 
& \Block{1-3}{\textbf{BEA60K}} & & & \Block{1-3}{\textbf{JFLEG}} & & & \Block{1-3}{\textbf{BEA60K}} &  & & \Block{1-3}{\textbf{JFLEG}} & & &\Block{1-3}{\textbf{BEA60K}} &  & & \Block{1-3}{\textbf{JFLEG}} \\
& \scriptsize{Pr.} & \scriptsize{Rec.} & \scriptsize{F1} & \scriptsize{Pr.} & \scriptsize{Rec.} & \scriptsize{F1} & \scriptsize{Pr.} & \scriptsize{Rec.} & \scriptsize{F1} & \scriptsize{Pr.} & \scriptsize{Rec.} & \scriptsize{F1} & \scriptsize{Pr.} & \scriptsize{Rec.} & \scriptsize{F1} & \scriptsize{Pr.} & \scriptsize{Rec.} & \scriptsize{F1}\\

\hline

\footnotesize
\Block[l]{}{\textbf{Full Prompt}} & & & & & & \\
%\multicolumn{1}{|r|}{Item3}
\Block[l]{}{\;\;\;\scriptsize{W/O Punctuation}} 

& \scriptsize{\textbf{66.9}} & \scriptsize{84.1} & \scriptsize{\textbf{74.5}} & \scriptsize{\textbf{77.8}} & \scriptsize{88.6} & \scriptsize{\textbf{82.9}}
& \scriptsize{\textbf{68.7}} & \scriptsize{85.3} & \scriptsize{\textbf{76.1}} & \scriptsize{\textbf{77.9}} & \scriptsize{88.3} & \scriptsize{\textbf{82.8}}
& \scriptsize{\textbf{67.7}} & \scriptsize{84.0} & \scriptsize{\textbf{75.0}} & \scriptsize{\textbf{76.8}} & \scriptsize{88.5} & \scriptsize{\textbf{82.2}} \\
\Block[l]{}{\;\;\;\scriptsize{With Punctuation}} 

& \scriptsize{57.1} & \scriptsize{83.5} & \scriptsize{67.8} & \scriptsize{73.3} & \scriptsize{87.9} & \scriptsize{80.0}
& \scriptsize{58.6} & \scriptsize{84.5} & \scriptsize{69.2} & \scriptsize{73.5} & \scriptsize{87.7} & \scriptsize{80.0}
& \scriptsize{57.6} & \scriptsize{83.3} & \scriptsize{68.1} & \scriptsize{72.7} & \scriptsize{87.9} & \scriptsize{79.6} \\
\footnotesize
\Block[l]{}{\textbf{Short Prompt}} & & & & & & \\
%\multicolumn{1}{|r|}{Item3}
\Block[l]{}{\;\;\;\scriptsize{W/O Punctuation}} 
& \scriptsize{38.7} & \scriptsize{\textbf{86.3}} & \scriptsize{53.5} & \scriptsize{43.5} & \scriptsize{\textbf{89.5}} & \scriptsize{58.6}
& \scriptsize{39.0} & \scriptsize{\textbf{85.5}} & \scriptsize{53.5} & \scriptsize{39.5} & \scriptsize{\textbf{90.3}} & \scriptsize{55.0}
& \scriptsize{38.6} & \scriptsize{\textbf{86.5}} & \scriptsize{53.4} & \scriptsize{40.1} & \scriptsize{\textbf{90.5}} & \scriptsize{55.6}\\
\Block[l]{}{\;\;\;\scriptsize{With Punctuation}} 
& \scriptsize{34.4} & \scriptsize{85.5} & \scriptsize{49.0} &  \scriptsize{41.9} & \scriptsize{89.0} & \scriptsize{57.0}
& \scriptsize{34.7} & \scriptsize{84.9} & \scriptsize{49.2} & \scriptsize{37.9} & \scriptsize{89.7} & \scriptsize{53.3}
& \scriptsize{34.7} & \scriptsize{85.9} & \scriptsize{49.4} & \scriptsize{38.6} & \scriptsize{90.0} & \scriptsize{54.0} \\

\footnotesize
\Block[l]{}{\textbf{Cut Prompt}} & & & & & & \\
%\multicolumn{1}{|r|}{Item3}
\Block[l]{}{\;\;\;\scriptsize{W/O Punctuation}} 
& \scriptsize{22.6} & \scriptsize{80.3} & \scriptsize{35.3} & \scriptsize{20.5} & \scriptsize{80.8} & \scriptsize{32.7}
& \scriptsize{22.7} & \scriptsize{80.2} & \scriptsize{35.4} & \scriptsize{21.5} & \scriptsize{83.7} & \scriptsize{34.3}
& \scriptsize{22.3} & \scriptsize{80.2} & \scriptsize{34.9} & \scriptsize{21.1} & \scriptsize{83.1} & \scriptsize{33.7} \\
\Block[l]{}{\;\;\;\scriptsize{With Punctuation}} 
& \scriptsize{20.6} & \scriptsize{79.6} & \scriptsize{32.8} & \scriptsize{19.9} & \scriptsize{79.9} & \scriptsize{31.9}
& \scriptsize{20.8} & \scriptsize{79.5} & \scriptsize{33.0} & \scriptsize{20.8} & \scriptsize{82.9} & \scriptsize{33.3} 
& \scriptsize{20.4} & \scriptsize{80.1} & \scriptsize{32.6} & \scriptsize{20.7} & \scriptsize{82.5} & \scriptsize{33.1}\\ \hline
\end{NiceTabular*}

\caption{OpenAI models' performance on SC tasks in English. $W/O Punctuation$ and $With Punctuation$ reflect absence and presence of punctuation in sentence respectively. Metrics are reported in format \textbf{Prec}ision, \textbf{Rec}all, \textbf{F1}-measure from ~\cite{sorokin2016spellrueval}.}
\label{tab:enprompts}
\end{table*}
\endgroup

\begin{table*}[htbp!]
\centering
\footnotesize
\begin{NiceTabular}{{c}|{c}{c}|{c}{c}|{c}{c}}%[hvlines]
\hline
\Block{2-1}{\textbf{Prompt}} & \Block{1-2}{\textbf{gpt-3.5-turbo-0301}} &  & \Block{1-2}{\textbf{gpt-4-0314}} &  & \Block{1-2}{\textbf{text-davinci-003}} \\ 
& \scriptsize{W/O Punctuation} & \scriptsize{With Punctuation} & \scriptsize{W/O Punctuation} & \scriptsize{With Punctuation} & \scriptsize{W/O Punctuation} & \scriptsize{With Punctuation}\\

\hline
\Block[l]{}{\textbf{Full Prompt}} & & & & & & \\
%\multicolumn{1}{|r|}{Item3}
\Block[l]{}{\;\;\;\scriptsize{RUSpellRU}} 
& \scriptsize{55.3} / \scriptsize{\textbf{75.8}} / \scriptsize{63.9}
& \scriptsize{\textbf{55.8}} / \scriptsize{75.3} / \scriptsize{\textbf{64.1}}
& \scriptsize{56.4} / \scriptsize{\textbf{76.2}} / \scriptsize{64.8}
& \scriptsize{\textbf{57.0}} / \scriptsize{75.9} / \scriptsize{\textbf{65.1}}
& \scriptsize{55.4} / \scriptsize{\textbf{75.8}} / \scriptsize{64.0}
& \scriptsize{\textbf{55.9}} / \scriptsize{75.3} / \scriptsize{\textbf{64.2}} \\

\Block[l]{}{\;\;\;\scriptsize{MultidomainGold}} 
& \scriptsize{30.8} / \scriptsize{70.9} / \scriptsize{43.0}
& \scriptsize{\textbf{33.8}} / \scriptsize{\textbf{72.1}} / \scriptsize{\textbf{46.0}}
& \scriptsize{31.0} / \scriptsize{72.0} / \scriptsize{43.3}
& \scriptsize{\textbf{34.0}} / \scriptsize{\textbf{73.2}} / \scriptsize{\textbf{
46.4}}
& \scriptsize{31.2} / \scriptsize{71.1} / \scriptsize{43.4}
& \scriptsize{\textbf{33.6}} / \scriptsize{\textbf{72.0}} / \scriptsize{\textbf{45.8}} \\

\Block[l]{}{\;\;\;\scriptsize{MedSpellChecker}} 
& \scriptsize{53.2} / \scriptsize{67.6} / \scriptsize{\textbf{59.6}}
& \scriptsize{\textbf{53.7}} / \scriptsize{66.1} / \scriptsize{59.3}
& \scriptsize{\textbf{54.2}} / \scriptsize{69.4} / \scriptsize{\textbf{60.9}}
& \scriptsize{\textbf{54.2}} / \scriptsize{67.7} / \scriptsize{60.2}
& \scriptsize{47.8} / \scriptsize{68.4} / \scriptsize{\textbf{56.3}}
& \scriptsize{\textbf{48.0}} / \scriptsize{66.4} / \scriptsize{55.7} \\

\Block[l]{}{\;\;\;\scriptsize{GitHubTypoCorpusRu}} 
& \scriptsize{\textbf{44.5}} / \scriptsize{\textbf{58.1}} / \scriptsize{\textbf{50.4}}
& \scriptsize{43.8} / \scriptsize{57.0} / \scriptsize{49.6}
& \scriptsize{\textbf{45.2}} / \scriptsize{\textbf{58.2}} / \scriptsize{\textbf{51.0}}
& \scriptsize{44.2} / \scriptsize{57.4} / \scriptsize{50.0}
& \scriptsize{\textbf{46.5}} / \scriptsize{\textbf{58.1}} / \scriptsize{\textbf{51.7}}
& \scriptsize{45.7} / \scriptsize{57.3} / \scriptsize{50.9} \\

\Block[l]{}{\textbf{Short Prompt}} & & & & & & \\
%\multicolumn{1}{|r|}{Item3}
\Block[l]{}{\;\;\;\scriptsize{RUSpellRU}} 
& \scriptsize{23.1} / \scriptsize{63.9} / \scriptsize{34.0}
& \scriptsize{23.8} / \scriptsize{63.5} / \scriptsize{34.7}
& \scriptsize{22.3} / \scriptsize{60.7} / \scriptsize{32.7}
& \scriptsize{23.2} / \scriptsize{60.5} / \scriptsize{33.6}
& \scriptsize{24.3} / \scriptsize{63.5} / \scriptsize{35.2}
& \scriptsize{25.2} / \scriptsize{63.6} / \scriptsize{36.1} \\

\Block[l]{}{\;\;\;\scriptsize{MultidomainGold}}
& \scriptsize{12.7} / \scriptsize{54.4} / \scriptsize{20.6}
& \scriptsize{15.0} / \scriptsize{55.8} / \scriptsize{23.6}
& \scriptsize{13.5} / \scriptsize{55.6} / \scriptsize{21.7}
& \scriptsize{15.4} / \scriptsize{55.9} / \scriptsize{24.1}
& \scriptsize{13.8} / \scriptsize{56.5} / \scriptsize{22.2}
& \scriptsize{16.1} / \scriptsize{57.7} / \scriptsize{25.2} \\

\Block[l]{}{\;\;\;\scriptsize{MedSpellChecker}} 
& \scriptsize{30.7} / \scriptsize{76.1} / \scriptsize{43.8}
& \scriptsize{29.2} / \scriptsize{\textbf{77.9}} / \scriptsize{42.5}
&\scriptsize{29.0} / \scriptsize{\textbf{78.6}} / \scriptsize{42.4}
& \scriptsize{30.6} / \scriptsize{76.9} / \scriptsize{43.8}
& \scriptsize{29.8} / \scriptsize{76.4} / \scriptsize{42.9}
& \scriptsize{28.4} / \scriptsize{\textbf{77.9}} / \scriptsize{41.7} \\

\Block[l]{}{\;\;\;\scriptsize{GitHubTypoCorpusRu}} 
& \scriptsize{18.4} / \scriptsize{45.8} / \scriptsize{26.3}
& \scriptsize{18.8} / \scriptsize{46.9} / \scriptsize{26.9}
&  \scriptsize{17.1} / \scriptsize{46.0} / \scriptsize{25.0}
& \scriptsize{17.7} / \scriptsize{47.1} / \scriptsize{25.7}
& \scriptsize{19.7} / \scriptsize{47.1} / \scriptsize{27.8}
& \scriptsize{20.1} / \scriptsize{47.1} / \scriptsize{28.2} \\

\Block[l]{}{\textbf{Cut Prompt}} & & & & & & \\
%\multicolumn{1}{|r|}{Item3}
\Block[l]{}{\;\;\;\scriptsize{RUSpellRU}} 
& \scriptsize{37.9} / \scriptsize{70.3} / \scriptsize{49.3}
& \scriptsize{38.8} / \scriptsize{70.1} / \scriptsize{50.0}
& \scriptsize{35.6} / \scriptsize{64.1} / \scriptsize{45.8}
& \scriptsize{36.4} / \scriptsize{64.0} / \scriptsize{46.4}
& \scriptsize{37.0} / \scriptsize{69.5} / \scriptsize{48.3}
& \scriptsize{37.9} / \scriptsize{69.4} / \scriptsize{49.0} \\

\Block[l]{}{\;\;\;\scriptsize{MultidomainGold}} 
& \scriptsize{7.2} / \scriptsize{46.4} / \scriptsize{12.5}
& \scriptsize{7.5} / \scriptsize{49.1} / \scriptsize{13.1}
& \scriptsize{10.5} / \scriptsize{62.1} / \scriptsize{18.0}
& \scriptsize{7.6} / \scriptsize{46.3} / \scriptsize{13.0}
& \scriptsize{10.6} / \scriptsize{60.6} / \scriptsize{18.0}
& \scriptsize{12.3} / \scriptsize{62.0} / \scriptsize{20.6} \\

\Block[l]{}{\;\;\;\scriptsize{MedSpellChecker}} 
& \scriptsize{5.5} / \scriptsize{52.2} / \scriptsize{10.0}
& \scriptsize{5.3} / \scriptsize{56.3} / \scriptsize{9.7}
& \scriptsize{4.7} / \scriptsize{49.7} / \scriptsize{8.6}
& \scriptsize{5.6} / \scriptsize{51.9} / \scriptsize{10.2}
& \scriptsize{5.9} / \scriptsize{59.9} / \scriptsize{10.8}
& \scriptsize{6.5} / \scriptsize{57.6} / \scriptsize{11.7} \\

\Block[l]{}{\;\;\;\scriptsize{GitHubTypoCorpusRu}} 
& \scriptsize{17.0} / \scriptsize{50.4} / \scriptsize{25.4}
& \scriptsize{17.2} / \scriptsize{50.3} / \scriptsize{25.7}
& \scriptsize{18.0} / \scriptsize{52.7} / \scriptsize{26.8}
& \scriptsize{18.4} / \scriptsize{53.5} / \scriptsize{27.4}
& \scriptsize{18.7} / \scriptsize{53.0} / \scriptsize{27.7}
& \scriptsize{18.6} / \scriptsize{53.3} / \scriptsize{27.6} \\ \hline

\end{NiceTabular}

\caption{OpenAI models' performance on SC task in Russian. $W/O Punctuation$ and $With Punctuation$ reflect absence and presence of punctuation in sentence respectively. Metrics are reported in format \textbf{Prec}ision, \textbf{Rec}all, \textbf{F1}-measure from ~\cite{sorokin2016spellrueval}.}
\label{tab:prompts}
\end{table*}

\subsection{Hyperparameters}
\label{sec:appendix:hyperparameters}

\begin{table*}[htbp!]  % \label{table_results}
\centering
\footnotesize
\begin{NiceTabular}{{c}|{c}{c}{c}{c}{c}}%[hvlines]
\hline
\Block{2-1}{\textbf{Model}} & \Block{1-5}{\textbf{Hyperparameters}} & & & & \\
& \scriptsize{learning rate} & \scriptsize{weight decay} & \scriptsize{warmup steps} & \scriptsize{batch size} & \scriptsize{epochs}\\
\hline
\Block[l]{}{\textbf{M2M100-1.2B}} & & & & & \\
%\multicolumn{1}{|r|}{Item3}
\Block[l]{}{\;\;\scriptsize{\underline{Fine-tuning}}} & & & & & \\
\Block[l]{}{\;\;\;\;\scriptsize{RUSpellRU}} 
& 8.62e-5 & 0.0288 & 5 & 16 & 7 \\
\Block[l]{}{\;\;\;\;\scriptsize{MultidomainGold}} 
& 4.96e-5 & 0.0135 & 5 & 16 & 8 \\
\Block[l]{}{\;\;\;\;\scriptsize{RUSpellRU+MDG}} 
& 6.48e-5 & 0.0416 & 10 & 16 & 7 \\
\Block[l]{}{\;\;\scriptsize{\underline{Pr. + Fine-tuning}}} & & & & & \\
\Block[l]{}{\;\;\;\;\scriptsize{RUSpellRU}} 
& 8.62e-5 & 0.0288 & 5 & 16 & 7 \\
\Block[l]{}{\;\;\;\;\scriptsize{MultidomainGold}} 
& 4.96e-5 & 0.0135 & 5 & 16 & 8 \\
\Block[l]{}{\;\;\;\;\scriptsize{RUSpellRU+MDG}} 
& 6.48e-5 & 0.0416 & 10 & 16 & 7 \\
\Block[l]{}{\;\;\scriptsize{\underline{Augmentex}}} & & & & & \\
\Block[l]{}{\;\;\;\;\scriptsize{RUSpellRU}} 
& 2e-5 & 0.01 & 0 & 8 & 7 \\
\Block[l]{}{\;\;\;\;\scriptsize{MultidomainGold}} 
& 2e-5 & 0.01 & 0 & 4 & 7 \\
\Block[l]{}{\;\;\;\;\scriptsize{RUSpellRU+MDG}} 
& 2e-5 & 0.01 & 0 & 4 & 7 \\
\Block[l]{}{\;\;\scriptsize{\underline{SBSC}}} & & & & & \\
\Block[l]{}{\;\;\;\;\scriptsize{RUSpellRU}} 
& 8.62e-5 & 0.0288 & 5 & 16 & 7 \\
\Block[l]{}{\;\;\;\;\scriptsize{MultidomainGold}} 
& 4.96e-5 & 0.0135 & 5 & 16 & 8 \\
\Block[l]{}{\;\;\;\;\scriptsize{RUSpellRU+MDG}} 
& 6.48e-5 & 0.0416 & 10 & 16 & 7 \\
\hline
\Block[l]{}{\textbf{M2M100-418M}} & & & & & \\
%\multicolumn{1}{|r|}{Item3}
\Block[l]{}{\;\;\scriptsize{\underline{Fine-tuning}}} & & & & & \\
\Block[l]{}{\;\;\;\;\scriptsize{RUSpellRU}} 
& 4.56e-5 & 0.0493 & 5 & 16 & 7 \\
\Block[l]{}{\;\;\;\;\scriptsize{MultidomainGold}} 
& 3.39e-5 & 0.0182 & 7 & 16 & 7 \\
\Block[l]{}{\;\;\;\;\scriptsize{RUSpellRU+MDG}} 
& 2.66e-5 & 0.0314 & 15 & 8 & 7 \\
\Block[l]{}{\;\;\scriptsize{\underline{Pr. + Fine-tuning}}} & & & & & \\
\Block[l]{}{\;\;\;\;\scriptsize{RUSpellRU}} 
& 4.56e-5 & 0.0493 & 5 & 16 & 7 \\
\Block[l]{}{\;\;\;\;\scriptsize{MultidomainGold}} 
& 3.39e-5 & 0.0182 & 7 & 16 & 7 \\
\Block[l]{}{\;\;\;\;\scriptsize{RUSpellRU+MDG}} 
& 2.66e-5 & 0.0314 & 15 & 8 & 7 \\
\Block[l]{}{\;\;\scriptsize{\underline{Augmentex}}} & & & & & \\
\Block[l]{}{\;\;\;\;\scriptsize{RUSpellRU}} 
& 2e-5 & 0.01 & 0 & 16 & 7 \\
\Block[l]{}{\;\;\;\;\scriptsize{MultidomainGold}} 
& 2e-5 & 0.01 & 0 & 8 & 7 \\
\Block[l]{}{\;\;\;\;\scriptsize{RUSpellRU+MDG}} 
& 2e-5 & 0.01 & 0 & 8 & 7 \\
\Block[l]{}{\;\;\scriptsize{\underline{SBSC}}} & & & & & \\
\Block[l]{}{\;\;\;\;\scriptsize{RUSpellRU}} 
& 4.56e-5 & 0.0493 & 5 & 16 & 7 \\
\Block[l]{}{\;\;\;\;\scriptsize{MultidomainGold}} 
& 3.39e-5 & 0.0182 & 7 & 16 & 7 \\
\Block[l]{}{\;\;\;\;\scriptsize{RUSpellRU+MDG}} 
& 2.66e-5 & 0.0314 & 15 & 8 & 7 \\
\hline
\Block[l]{}{\textbf{FredT5-large}} & & & & & \\
%\multicolumn{1}{|r|}{Item3}
\Block[l]{}{\;\;\scriptsize{\underline{Fine-tuning}}} & & & & & \\
\Block[l]{}{\;\;\;\;\scriptsize{RUSpellRU}} 
& 2e-4 & 0.01 & 0 & 8 & 10 \\
\Block[l]{}{\;\;\;\;\scriptsize{MultidomainGold}} 
& 2e-4 & 0.01 & 0 & 8 & 10 \\
\Block[l]{}{\;\;\;\;\scriptsize{RUSpellRU+MDG}} 
& 2e-4 & 0.01 & 0 & 8 & 8 \\
\Block[l]{}{\;\;\scriptsize{\underline{Pr. + Fine-tuning}}} & & & & & \\
\Block[l]{}{\;\;\;\;\scriptsize{RUSpellRU}} 
& 2e-4 & 0.01 & 0 & 8 & 10 \\
\Block[l]{}{\;\;\;\;\scriptsize{MultidomainGold}} 
& 2e-4 & 0.01 & 0 & 8 & 10 \\
\Block[l]{}{\;\;\;\;\scriptsize{RUSpellRU+MDG}} 
& 2e-4 & 0.01 & 0 & 8 & 8 \\
\Block[l]{}{\;\;\scriptsize{\underline{Augmentex}}} & & & & & \\
\Block[l]{}{\;\;\;\;\scriptsize{RUSpellRU}} 
& 2e-4 & 0.01 & 0 & 8 & 10 \\
\Block[l]{}{\;\;\;\;\scriptsize{MultidomainGold}} 
& 2e-4 & 0.01 & 0 & 8 & 10 \\
\Block[l]{}{\;\;\;\;\scriptsize{RUSpellRU+MDG}} 
& 2e-4 & 0.01 & 0 & 8 & 8 \\
\Block[l]{}{\;\;\scriptsize{\underline{SBSC}}} & & & & & \\
\Block[l]{}{\;\;\;\;\scriptsize{RUSpellRU}} 
& 2e-4 & 0.01 & 0 & 8 & 10 \\
\Block[l]{}{\;\;\;\;\scriptsize{MultidomainGold}} 
& 2e-4 & 0.01 & 0 & 8 & 10 \\
\Block[l]{}{\;\;\;\;\scriptsize{RUSpellRU+MDG}} 
& 2e-4 & 0.01 & 0 & 8 & 8 \\ \hline

\end{NiceTabular}

\caption{The hyperparameters of models' configurations (pre-trained, fine-tuning, augmentation).}
\label{tab:hyperparameters}
\end{table*}

\end{document}